\documentclass[runningheads]{llncs}
\PassOptionsToPackage{table}{xcolor}

\usepackage{accv}

\usepackage{accvabbrv}

\usepackage{graphicx}
\usepackage{booktabs}

\usepackage[accsupp]{axessibility}  %

\newcommand{\red}[1]{{\color{red}#1}}
\newcommand{\blue}[1]{{\color{blue}#1}}

\newcommand{\best}[1]{{\textbf{#1}}}

\usepackage{tikz}
\usetikzlibrary{patterns}
\newcommand{\hatchbox}{\tikz{\filldraw[pattern=north east lines] (0,0) rectangle (0.4,0.2);}}
\newcommand{\solidbox}{\tikz{\filldraw[gray!50] (0,0) rectangle (0.4,0.2);}}
\usepackage{lipsum}
\usepackage{array}
\usepackage{times}
\usepackage{makecell}
\usepackage{epsfig}
\usepackage{graphicx}
\usepackage{float}
\usepackage{wrapfig}
\usepackage{amsmath,amssymb}
\usepackage{algorithm,algorithmicx,algpseudocode}
\usepackage{bm,xspace}
\usepackage{comment}
\usepackage{multirow}
\usepackage{balance}
\usepackage{url}
\usepackage{booktabs}
\usepackage{etoolbox,siunitx}
\usepackage{calc}
\usepackage{pifont,hologo}
\usepackage{color}
\usepackage{glossaries}
\usepackage{xcolor}
\usepackage{adjustbox}
\usepackage{amsmath}
\usepackage{enumitem}
\usepackage{bbding}  %
\usepackage[normalem]{ulem}  %
\usepackage{contour}
\usepackage{soul}%
\usepackage{tocloft} %
\usepackage{etoc} %
\PassOptionsToPackage{square,numbers,comma,sort}{natbib}

\definecolor{blue}{HTML}{004bb3}

\def\@onedot{\ifx\@let@token.\else.\null\fi\xspace}

\DeclareMathAlphabet{\mathsfit}{\encodingdefault}{\sfdefault}{m}{sl}
\SetMathAlphabet{\mathsfit}{bold}{\encodingdefault}{\sfdefault}{bx}{n}

\newcommand{\E}{\mathbb{E}}

\newcommand{\R}{\mathbb{R}}

\let\originalleft\left
\let\originalright\right
\renewcommand{\left}{\mathopen{}\mathclose\bgroup\originalleft}
\renewcommand{\right}{\aftergroup\egroup\originalright}

\definecolor{codegreen}{rgb}{0,0.6,0} %

\newacronym{vfm}{VFM}{Vision Foundation Model}
\newacronym{VFMs}{VFMs}{Vision Foundation Models}
\newacronym{fsad}{FSAD}{Few-shot anomaly detection}

\newacronym{clip}{CLIP}{Contrastive Language–Image Pretraining}
\newacronym{as}{AS}{anomaly segmentation}
\newacronym{mad}{MAD}{Medical anomaly detection}

\usepackage[pagebackref,breaklinks,colorlinks,citecolor=accvblue]{hyperref}

\usepackage{orcidlink}

\makeatletter
\g@addto@macro\UrlBreaks{%
  \do\a\do\b\do\c\do\d\do\e\do\f\do\g\do\h\do\i\do\j\do\k\do\l\do\m%
  \do\n\do\o\do\p\do\q\do\r\do\s\do\t\do\u\do\v\do\w\do\x\do\y\do\z%
  \do\A\do\B\do\C\do\D\do\E\do\F\do\G\do\H\do\I\do\J\do\K\do\L\do\M%
  \do\N\do\O\do\P\do\Q\do\R\do\S\do\T\do\U\do\V\do\W\do\X\do\Y\do\Z%
  \do\0\do\1\do\2\do\3\do\4\do\5\do\6\do\7\do\8\do\9\do\-\do\_%
}
\makeatother

\begin{document}

\title{ProX-MAD:\\Aggregating Spatial Context into Representative Proxies\\for Few-shot Medical Anomaly Detection\vspace{-0.5em}}

\titlerunning{ProX-MAD: Proxy-based Few-shot Medical Anomaly Detection}

\author{Seunghyun Oh\orcidlink{0009-0007-4660-5725} \and
Seongsu Lee\orcidlink{0009-0005-0098-3179} \and
Seunghye Chae\orcidlink{0009-0008-4197-5839} \and
Youngmin Ro\textsuperscript{*}\orcidlink{0000-0002-0661-9192}}

\authorrunning{Oh et al.}

\institute{Machine Intelligence Laboratory, \\ Dept. of Artificial Intelligence, University of Seoul, Korea\\
\email{\{osh1795, ssulee0206, tmdhey, youngmin.ro\}@uos.ac.kr}}

\maketitle
\begingroup
\renewcommand{\thefootnote}{*}
\footnotetext{Corresponding author.}
\endgroup

\vspace{-1.0em}
\begin{abstract}
Few-shot medical anomaly detection (MAD) has largely followed a dataset-specific "one-for-one" paradigm, leaving unified "one-for-many" modeling underexplored. Naively extending existing methods to this setting causes substantial performance degradation, especially at the pixel level. We attribute this to their detection mechanisms: a shared decision boundary becomes unreliable across heterogeneous lesion definitions, whereas a normal-only memory bank cannot adequately cover diverse normal distributions with few references. We propose \textbf{ProX-MAD}, a proxy-based framework that aggregates spatial context into compact, learnable, state-specific proxies for unified few-shot MAD. Its Masked Proxy Infusion Attention (MPIA) module cross-attends proxy tokens to reference patch features and applies proxy-wise masking to retain high-confidence correspondences, mitigating proxy collapse and encouraging diverse proxies that summarize both normal and abnormal references. Across six medical datasets, ProX-MAD exhibits only a marginal performance drop from one-for-one to one-for-many training. Despite being trained only once as a unified model, it also outperforms prior state-of-the-art methods on average even when they use their more favorable one-for-one protocol, demonstrating the practicality of unified multi-dataset MAD.
Code is available at \url{https://github.com/SeungHyun99/ProX-MAD}.

\keywords{Medical Anomaly Detection \and Few-shot Learning \and Unified Medical Anomaly Detection \and Vision Foundation Models}
\end{abstract}

\section{Introduction}
\label{sec:intro}
\glsresetall

\begin{figure*}[!t]%
  \centering
  \includegraphics[width=\textwidth]{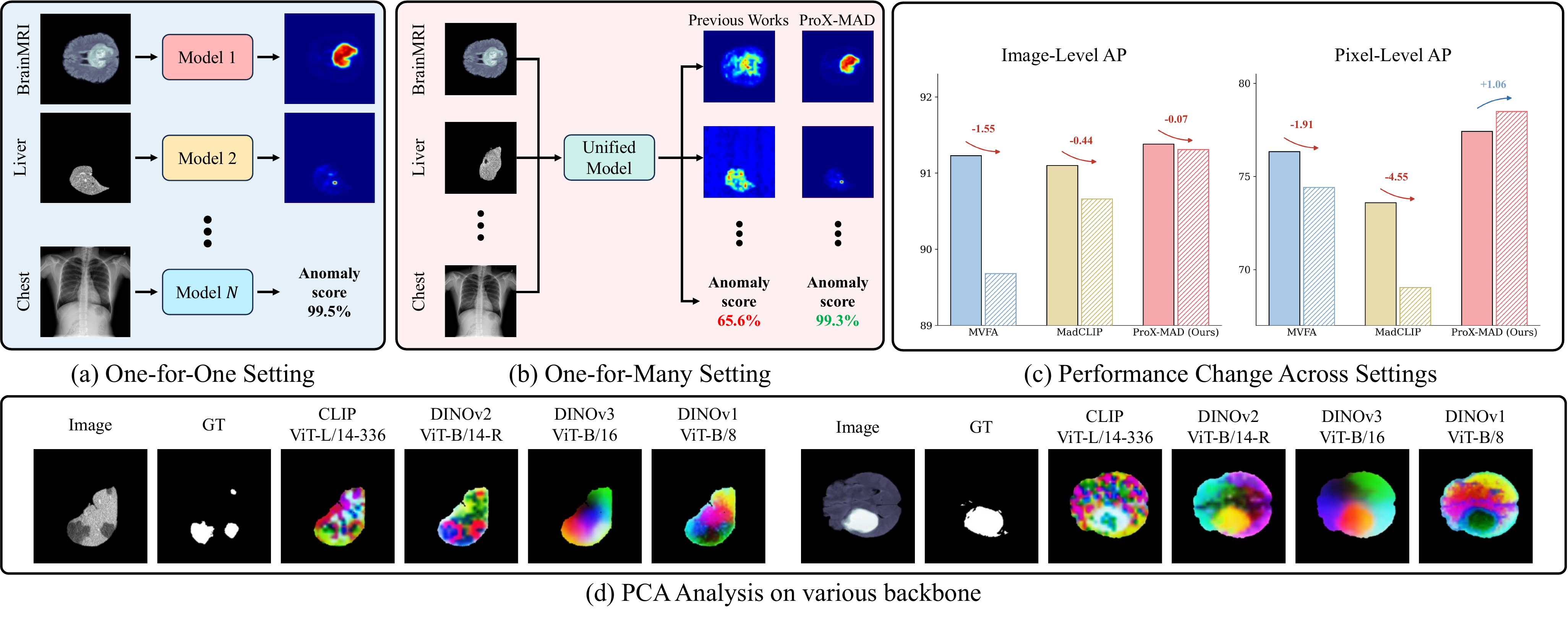}
  \vspace{-5mm}
  \caption{(a) \textbf{One-for-one setting}, where each model is trained on a single dataset. (b) \textbf{One-for-many setting}, where a single model is jointly trained on multiple datasets. (c) \textbf{Performance comparison} under the 2-shot setting. We report image- and pixel-level AP under both settings. Solid bars(\protect\solidbox) denote one-for-one results, while hatched bars(\protect\hatchbox) denote one-for-many results. (d) \textbf{PCA visualization} of spatial features from various vision foundation models.
}
  \label{fig:teaser}
  \vspace{-5mm}
\end{figure*}

Medical Anomaly Detection (MAD) facilitates early diagnosis and minimizes clinical risks by identifying and localizing pathological findings in medical imaging~\cite{ai_in_health}. Due to the difficulty of acquiring large-scale annotated data, recent studies in MAD have increasingly adopted few-shot learning approaches.
Few-shot MAD methods~\cite{mvfa,madclip} have largely followed a one-for-one setting, where a separate model is trained for each dataset or clinical domain. Although this task-specific specialization yields strong performance, training a single model once and applying it across heterogeneous modalities without additional fine-tuning—a one-for-many setting (\cref{fig:teaser}(b))—remains largely unexplored. Such unification eliminates per-dataset retraining and model storage, but requires a single model to reconcile different anatomical structures, imaging physics, and lesion definitions. Unified one-for-many modeling has been studied in general-domain anomaly detection~\cite{uni_uniad,uni_dinomaly,uni_mambdaad,uni_moead,uni_uninet}, but these methods target natural images with substantially more training data. Extending this paradigm to heterogeneous medical domains under few-shot supervision is the gap addressed in this work.

Existing few-shot MAD methods~\cite{mvfa, madclip} are trained under a few-shot supervised setting, using only a handful of annotated normal and abnormal samples per dataset. When these methods are instead trained jointly across all datasets under the one-for-many setting, their performance drops substantially, most severely at the pixel level, compared to their dedicated, one-for-one counterparts, as shown in \cref{fig:teaser}(c). We attribute this degradation to two requirements that a one-for-many model must satisfy. First, the underlying visual representation must separate normal and abnormal semantics consistently across organs and modalities. Second, the detection mechanism itself must remain consistent under domain heterogeneity while scaling gracefully as the number of domains and reference samples grows. As we show next, existing methods satisfy neither requirement. Crucially, while a spatially discriminative representation is a prerequisite, our experiments (\cref{sec:exps}) show that it alone does not close the one-for-many gap. The core difficulty lies in the detection mechanism, which is the main focus of our design.

To determine whether existing representations satisfy this first requirement, we examine the patch-level features on which prior few-shot MAD methods operate. Most of these methods are built on CLIP~\cite{clip}, a backbone pretrained for image-text alignment rather than dense spatial discrimination. We therefore conduct a PCA analysis across several vision foundation models (VFMs), including CLIP and three DINO variants~\cite{dino,vit_reg,dinov3}, to visualize how well their patch-level features separate normal from abnormal tissue (\cref{fig:teaser}(d)). CLIP features turn out to be spatially diffuse, with normal and abnormal regions blending together rather than forming distinguishable clusters. DINO features separate the two more cleanly, and between the three variants, the smaller patch size of DINOv1 (ViT-B/8) yields sharper, more fine-grained boundaries than DINOv2, which is particularly important for delineating small or irregularly shaped medical lesions.

The second requirement concerns the design of the detection mechanism itself, and existing methods fall into two paradigms—decision-boundary and memory-bank—each of which breaks down in a different way under domain heterogeneity. MadCLIP instantiates the decision-boundary approach, adapting CLIP's visual and textual representations to separate normal from abnormal patterns. While a handful of samples suffices to fit such a boundary within a single dataset, jointly modeling heterogeneous datasets makes this considerably harder: lesion definitions and distributional characteristics vary across domains, hindering the formation of consistent decision regions and often yielding blurred or domain-biased boundaries.

MVFA instead relies on a memory bank of normal patch-level features, against which query features are compared at inference. This design fails differently under domain heterogeneity. Because lesion size and shape vary substantially across datasets, a bank built from only a few normal references per domain cannot adequately represent each dataset's normal distribution, and it must grow with the number of shots and domains, increasing memory cost accordingly. Moreover, even when abnormal samples are available, such methods struggle to exploit them: a single abnormal patch typically mixes normal and pathological signal, so storing it directly yields ambiguous, unreliable features rather than clean pathological representations.

To address these challenges, we introduce \textbf{ProX-MAD}, a Proxy-based Medical Anomaly Detection framework. Motivated by the analysis above, ProX-MAD builds on the fine-grained spatial features of a DINO backbone and represents diverse pathological and healthy patterns through a set of compact, learnable tokens. Central to our approach is the Masked Proxy Infusion Attention (MPIA) module, a cross-attention mechanism designed to enrich these proxies with spatial context from reference images. Specifically, MPIA facilitates state-conditioned infusion, where proxies are selectively updated with features from their corresponding diagnostic states. To prevent representational redundancy, we apply a masking mechanism to the attention maps, encouraging each proxy to specialize in distinct semantic attributes. Finally, by comparing these specialized, infused proxies with query image features, ProX-MAD generates dense anomaly maps for precise localization.

Our contributions are as follows:
\begin{itemize}
    \item We propose \textbf{ProX-MAD}, a proxy-based framework that performs unified few-shot medical anomaly detection by representing normal and abnormal states with compact state-specific proxies.
    \item We introduce \textbf{Masked Proxy Infusion Attention} (MPIA), which injects state-matched spatial cues from reference images into each proxy and encourages diverse, complementary representations via masked cross-attention.
    \item We demonstrate that ProX-MAD maintains robust performance in the unified one-for-many setting, exhibiting only a marginal performance drop when scaling from dataset-specific training, while surpassing prior state-of-the-art few-shot MAD methods on average.
\end{itemize}
\section{Related Works}

\textbf{Vision Language and Vision Foundation Models.}
Vision-language models (VLMs) and vision foundation models (VFMs) have been widely adopted across various downstream vision tasks due to their strong representation learning capabilities. Among VLMs, CLIP~\cite{clip} is one of the most representative models. CLIP is trained using contrastive learning on large-scale image-text pairs, aligning visual and textual representations within a shared embedding space. Through language supervision, CLIP demonstrates strong generalization ability and achieves impressive performance on a wide range of visual recognition tasks, particularly in zero-shot settings~\cite{maskclip_tf, winclip, cocoop,aprilgan, anomalyclip}. However, CLIP is primarily trained with image-level supervision, which limits its ability to capture fine-grained spatial information~\cite{clip_dinoiser,clearclip,clip_surgery}. In addition, since CLIP is pretrained on natural image datasets, a substantial domain gap exists between its training data and medical images, often resulting in degraded performance in medical imaging applications~\cite{clip_mad_bad1,medclip,biomedclip}.

In contrast, self-supervised vision foundation models~\cite{dino,dinov2,vit_reg,dinov3,mae} learn visual representations directly from images without language supervision. By exploiting the intrinsic visual structure and semantic consistency of the data, these models produce rich and spatially informative feature representations. Consequently, self-supervised VFMs have demonstrated superior performance in dense prediction tasks~\cite{rein, soma,anomolydino}, such as segmentation and localization, where fine-grained spatial understanding is essential.

\noindent \textbf{Medical Anomaly Detection.}
Motivated by the difficulty of acquiring large-scale annotated data in clinical practice, medical anomaly detection has progressively shifted from training on large collections of normal images~\cite{anogan, f-anogan} toward few-shot settings that require only a handful of reference samples~\cite{mediclip}. More recently, few-shot MAD has advanced further by incorporating a small number of annotated abnormal images alongside normal references, giving rise to few-shot supervised methods such as MVFA~\cite{mvfa} and MadCLIP~\cite{madclip}. MVFA adapts a pretrained CLIP visual encoder with lightweight adapters and constructs a memory bank from the few-shot normal reference features for nearest-neighbor comparison. MadCLIP instead employs dual-branch adapters with learnable text prompts to contrast normal and abnormal representations. Unified one-for-many modeling has been actively studied in general-domain anomaly detection~\cite{uni_dinomaly, uni_mambdaad, uni_moead, uni_uniad,uni_uninet}, where a single model is trained across multiple object or texture classes. In contrast, few-shot MAD has remained confined almost entirely to the one-for-one setting, and its extension to this unified paradigm has received little attention. In this work, we introduce learnable proxies that scale gracefully across datasets and shots, capturing the diverse normal and abnormal patterns that characterize different medical modalities within a single unified model.
\section{Methods}
\label{sec:method}
\glsresetall

\subsection{Problem Formulation}
We aim to develop a general framework for few-shot medical anomaly detection in a one-for-many setting. To train a unified model, we use a combined training collection $\mathcal{D}^{\mathrm{train}}=\bigcup_{i\in\Omega}\mathcal{D}^{\mathrm{train}}_i$, where $\Omega$ is the index set over the target datasets $\{\mathcal{D}_i\}_{i\in\Omega}$.
For each dataset $\mathcal{D}^{\mathrm{train}}_i$, we construct a few-shot training set by sampling $k$ normal and $k$ anomalous samples, with $k\in\{1,2,4\}$. Each sample is represented as a triplet $(x, y, \mathcal{G})$, comprising an input image $x \in \mathbb{R}^{H \times W \times 3}$, a binary label $y \in \{0, 1\}$, and a spatial ground-truth mask $\mathcal{G} \in \{0, 1\}^{H \times W}$. In this convention, $0$ and $1$ denote normal and abnormal (pathological) states, respectively. Pixel-level masks $\mathcal{G}$ are available only for datasets with segmentation annotations and are omitted otherwise. Following training on $\mathcal{D}^{\mathrm{train}}$, the model performs anomaly classification~(AC) and anomaly segmentation~(AS) on each $\mathcal{D}^{\mathrm{test}}_i$ without further training.

\subsection{Overview}
\label{sec:overview}

\begin{figure*}[!t]%
  \centering
  \includegraphics[width=0.97\textwidth]{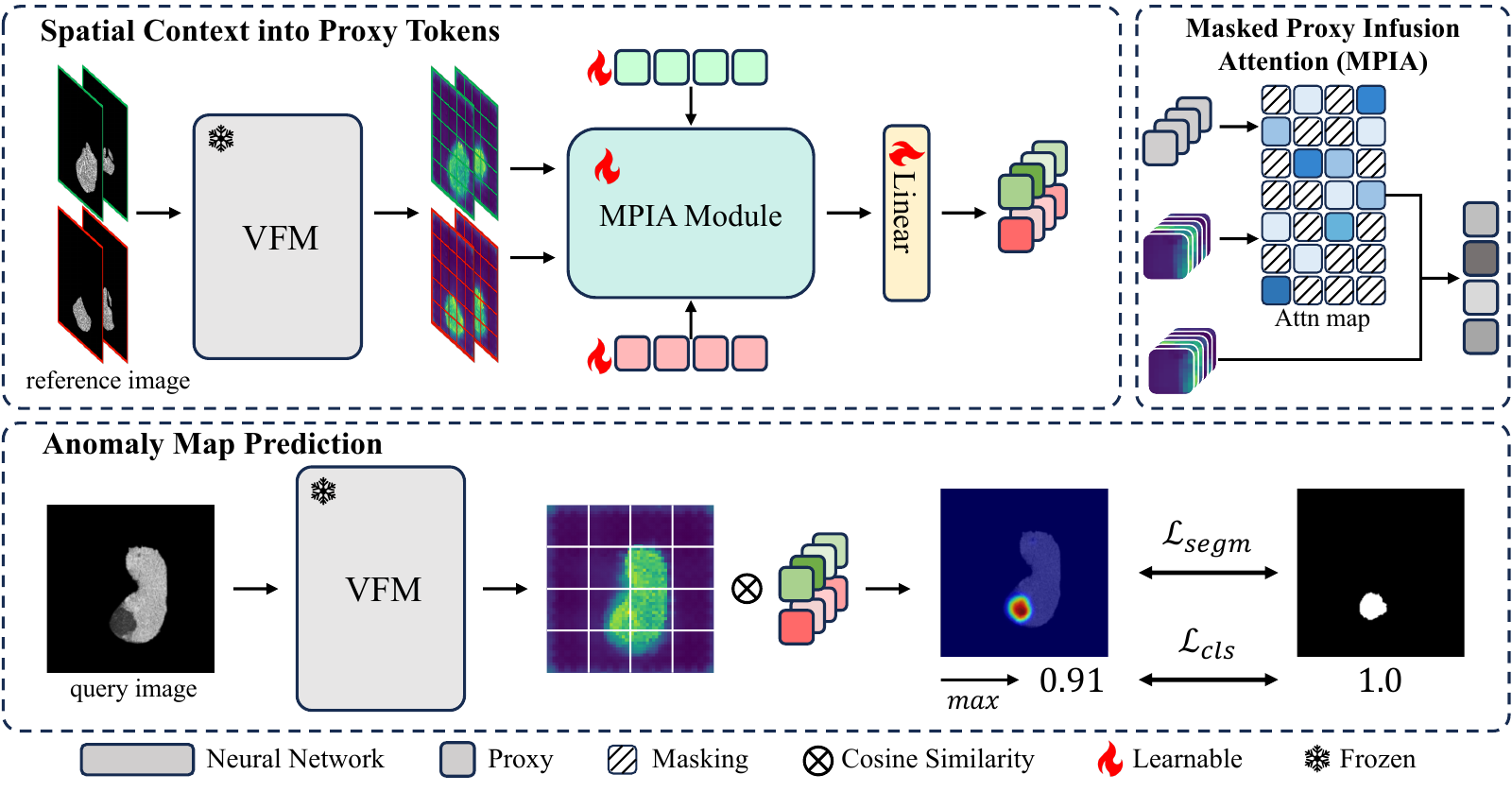}
  \caption{\textbf{Overview of the ProX-MAD pipeline.} Reference features are infused into learnable proxy tokens via Masked Proxy Infusion Attention (MPIA), and the anomaly map is obtained by comparing query features against the infused proxies.
}
  \label{fig:overview}
  \vspace{-3mm}
\end{figure*}
\cref{fig:overview} illustrates the overall pipeline of ProX-MAD. For each state, normal and abnormal, we prepare a set of learnable proxy tokens that represent that state. The proxy tokens of both states are passed through a shared cross-attention-based Masked Proxy Infusion Attention (MPIA) module, which infuses each proxy with spatial context from reference images of its corresponding state. The resulting infused proxies are then compared against the patch features of a query image to produce an anomaly map, and the maximum value of this map is taken as the image-level anomaly score. We apply this pipeline at multiple intermediate layers of the backbone; for notational brevity, we omit the explicit layer index $l$ throughout.

\subsection{Aggregating Spatial Context into Representative Proxies}
\label{sec:method_proxmad}

\noindent \textbf{Representative proxy tokens.}
In few-shot medical anomaly detection, the normal reference set is extremely small and may not reflect the full distribution of normal anatomy. As a result, directly storing all spatial tokens from these references in a memory bank can be suboptimal and overly dependent on the particular samples. To address this, ProX-MAD employs a set of $N$ learnable proxy tokens $P^{s}\in\mathbb{R}^{N\times d}$ for each state $s\in\{n, a\}$ (normal and abnormal), where $d$ is the backbone feature dimension. Each individual proxy is optimized to specialize in a distinct sub-pattern within its respective state, where the number $N$ dictates the granularity for capturing intra-class diversity.

\noindent\textbf{Masked Proxy Infusion Attention (MPIA).}
To improve state-wise representativeness, we infuse reference information into the proxy tokens via Masked Proxy Infusion Attention (MPIA), a single-head cross-attention module that refines each proxy based on reference features from the same state. Within each selected layer, the normal and abnormal proxies are processed by a single shared MPIA module; the two states differ only in their proxy tokens (queries) and reference features (keys and values), not in the module parameters.
For a given state $s\in\{n, a\}$, the proxy tokens~$P^s$ are utilized as queries, while spatial tokens~$F^s_{ref}\in\mathbb{R}^{n_r\times d}$ extracted from reference images of the same state are utilized as keys and values, where $n_r$ is the number of reference spatial tokens.
Using learnable projections, we form the query, key, and value as
\begin{equation}
q = \mathrm{Proj}_q(P^s),~~~ k = \mathrm{Proj}_k(F^s_{ref}),~~~ v = \mathrm{Proj}_v(F^s_{ref}).
\end{equation} 
We then compute the scaled attention score matrix $A = qk^\top/\sqrt{d}\in\mathbb{R}^{N\times n_r}$, which measures the relevance between each proxy token and all reference spatial tokens. Since not all reference tokens are informative for a given proxy, we sparsify $A$ with a proxy-wise mask:
\begin{equation}
\label{eq:mpia_mask}
\tilde{A}_{i,j} =
\begin{cases}
A_{i,j}, & A_{i,j} \ge \tau_i, \\
-\infty, & \text{otherwise},
\end{cases}
\qquad
\tau_i = \mathrm{Quantile}_{\alpha}\!\left(A_{i,:}\right),
\end{equation}
which sets the lowest $\alpha$ fraction of attention scores in each row $A_{i,:}$ to $-\infty$, where $\alpha \in (0,1)$ denotes the masking ratio. A larger $\alpha$ therefore keeps fewer, more strongly supported reference tokens per proxy.
After softmax, weakly related tokens receive near-zero weights, preventing proxies from aggregating noisy evidence and encouraging different proxies to specialize in distinct, highly supported patterns. We obtain the proxy updates by weighting the values with the masked attention map and applying an output projection:
\begin{equation}
    {P}^{s\prime} = \mathrm{Proj}_o\!\left(\mathrm{Softmax}(\tilde{A})\, v\right).
\end{equation}

Since ${P}^{s\prime}$ originates from learnable tokens (randomly initialized and optimized) whereas the query features $F_{q}\in\mathbb{R}^{hw\times d}$ are extracted from a frozen backbone, the two representations can follow mismatched distributions. To mitigate this gap, we apply a lightweight linear projection $\phi(\cdot)$ to align the proxy tokens into a shared space, yielding the aligned infused proxies $\tilde{P}^{s} = \phi(P^{s\prime})$. Within each layer, the MPIA projections $\mathrm{Proj}_{q},\mathrm{Proj}_{k},\mathrm{Proj}_{v},\mathrm{Proj}_{o}$ and the alignment projection $\phi$ are shared by the normal and abnormal states; each selected layer instantiates its own MPIA module and its own $\phi$. Consequently, the only state-specific parameters are the learnable proxy tokens ($N$ per state, $2N$ per layer). The similarity-based anomaly map obtained from $\tilde{P}^{s}$ and the query features is detailed in \cref{sec:train_infer}.

\subsection{Training and Inference}
\label{sec:train_infer}

\noindent\textbf{Reference sampling.}
During training, for each query image we randomly sample a single normal reference $x^-$ and a single abnormal reference $x^+$ from the same dataset's $k$-shot pool, resampling them at every iteration. Although
$k$ references per state are available, infusing all of them at once could cause the model to overfit to a fixed reference configuration. Instead, the stochastic single-reference pairing acts as a form of reference-level augmentation: it exposes each proxy to varied reference appearances and encourages the model to generalize robustly across heterogeneous acquisition conditions. At inference we lift this restriction and use all $k$ references per state.

\noindent\textbf{Training.}
We predict anomaly maps by computing the similarity between the aligned infused proxies $\tilde{P}^{s}$ and the query features $F_{q}$, and optimize the model by comparing the predicted maps against the ground truth via a loss function. Specifically, the similarity score between the $i$-th proxy and the feature vector at the $j$-th spatial location of the query feature map is computed as

\begin{equation}
S^{a}_{ij}
=
\frac{
(\tilde{P}^{a}_{i})^\top F_{q,j}
}{
\|\tilde{P}^{a}_{i}\|_2 \,
\|F_{q,j}\|_2
},
\qquad
S^{n}_{ij}
=
\frac{
(\tilde{P}^{n}_{i})^\top F_{q,j}
}{
\|\tilde{P}^{n}_{i}\|_2 \,
\|F_{q,j}\|_2
}.
\end{equation}

where $i \in \{1,\ldots,N\}$ indexes the proxy,
$j \in \{1,\ldots,hw\}$ indexes the spatial location in the query feature map and $S^{n}, S^{a} \in \mathbb{R}^{N \times h \times w}$. For each state $s$, we aggregate the proxy dimension by taking the maximum response over proxies, yielding a state-wise map $M^{s}\in\mathbb{R}^{h\times w}$. We then apply a softmax over the two states (normal vs abnormal) and bilinearly upsample the abnormal probability map using \textit{Up} to obtain the anomaly map:
\begin{equation}
    \hat{M}=\textit{Up}\!\left(\big[\mathrm{Softmax}\!\left([M^{n},\, M^{a}]\right)\big]_{a}\right),
\end{equation}
where $[\,\cdot\,]_{a}$ selects the abnormal channel of the two-state softmax.

When pixel-level ground-truth masks are available, we supervise $\hat{M}$ with a segmentation objective:
\begin{equation}
    L_{\mathrm{segm}} = \lambda_{1}\,\mathrm{dice}(\hat{M}, \mathcal{G}) + \lambda_{2}\,\mathrm{focal}(\hat{M}, \mathcal{G}),
\end{equation}
where $\lambda_{1}$ and $\lambda_{2}$ balance the relative contribution of the Dice and focal losses~\cite{dice, focal}, respectively. We omit $\mathcal{L}_{\mathrm{segm}}$ when pixel-level annotations are unavailable. We obtain an image-level anomaly score as $\hat{y} = \max(\hat{M})$ and optimize it with a binary cross-entropy loss $L_{\mathrm{BCE}} = \mathrm{BCE}(\hat{y}, y)$. Both $L_{\mathrm{segm}}$ and $L_{\mathrm{BCE}}$ are computed independently at each selected layer, and the total loss $L_{total}$ is obtained by summing these per-layer losses across all selected layers:
\begin{equation}
    L_{total} = \sum_{\text{selected layers}} \left( L_{\mathrm{segm}} + \lambda_{3} L_{\mathrm{BCE}} \right).
\end{equation}
where $\lambda_{3}$ balances the relative contribution of the classification loss to the segmentation loss.

\noindent \textbf{Inference.}
At inference time, we use all $k$ normal and $k$ abnormal reference images from the target dataset.
We pre-compute the state-specific infused proxies $\tilde{P}^{s}$ once per dataset and cache them for efficient evaluation.
Given a test image $x_{\text{test}}$, we load the cached $\tilde{P}^{s}$ and compute the anomaly map and the image-level anomaly score at each selected layer.
We obtain the final anomaly map and anomaly score by averaging the layer-wise outputs.

\noindent\textbf{Complexity.}
Each state is summarized by $N$ proxy tokens, where $N$ serves as a capacity hyperparameter selected for the target training setting and remains fixed thereafter. For a given $N$, the cached representation and query-time matching cost scale as $O(N)$ and are independent of the shot count $k$. In contrast, a memory-bank baseline stores $O(k\cdot hw)$ reference tokens per dataset, so its memory footprint and matching cost grow with the number of reference samples.

\section{Experiments}
\label{sec:exps}
\glsresetall

\providecommand{\best}[1]{\textbf{#1}}
\providecommand{\ph}{\textcolor{gray}{--}}   %
\begin{table*}[t]
\centering
\caption{\textbf{Performance comparison under the one-for-one setting} for $k$ = 1, 2, and 4 shots. All methods (MVFA, MadCLIP, ProX-MAD) are evaluated with identical reference images per $k$-shot setting to ensure fair comparison. Image- and pixel-level results are averaged over three independently sampled few-shot reference sets (mean$_{\pm\text{std}}$); for the AVG row, both performance metrics and standard deviations are macro-averaged across datasets. Best results are in \best{bold}.}
\label{tab:main_sep}
\small
\setlength{\tabcolsep}{3.5pt}
\renewcommand{\arraystretch}{1.15}
\setlength{\aboverulesep}{0pt}\setlength{\belowrulesep}{0pt}\setlength{\cmidrulesep}{0.5pt}
\resizebox{\linewidth}{!}{%

\begin{tabular}{ll|cccccc|cccccc}
\toprule
\multirow{3}{*}{$k$} & \multirow{3}{*}{Dataset}
 & \multicolumn{6}{c|}{Image-level} & \multicolumn{6}{c}{Pixel-level} \\
\cmidrule(lr){3-8}\cmidrule(lr){9-14}
 & & \multicolumn{2}{c}{MVFA~\cite{mvfa}} & \multicolumn{2}{c}{MadCLIP~\cite{madclip}} & \multicolumn{2}{c|}{\textbf{ProX-MAD}} & \multicolumn{2}{c}{MVFA~\cite{mvfa}} & \multicolumn{2}{c}{MadCLIP~\cite{madclip}} & \multicolumn{2}{c}{\textbf{ProX-MAD}} \\
\cmidrule(lr){3-4}\cmidrule(lr){5-6}\cmidrule(lr){7-8}
\cmidrule(lr){9-10}\cmidrule(lr){11-12}\cmidrule(lr){13-14}
 & & AUROC & AP & AUROC & AP & AUROC & AP & AUROC & AP & AUROC & AP & AUROC & AP \\
\midrule
\multirow{7}{*}{\rotatebox{90}{1\,-\,shot}} & His & 70.21$_{\pm9.4}$ & \best{68.78$_{\pm2.7}$} & 69.13$_{\pm7.7}$ & 68.4$_{\pm1.2}$ & \best{71.43$_{\pm4.8}$} & 67.88$_{\pm3.3}$ & - & - & - & - & - & - \\
 & Chest & 64.74$_{\pm9.2}$ & 97.21$_{\pm0.9}$ & \best{68.49$_{\pm7.1}$} & 97.53$_{\pm0.8}$ & 68.45$_{\pm12.0}$ & \best{97.55$_{\pm1.2}$} & - & - & - & - & - & - \\
 & OCT17 & 81.41$_{\pm18.4}$ & 92.58$_{\pm7.7}$ & 78.94$_{\pm20.3}$ & 91.33$_{\pm8.4}$ & \best{93.68$_{\pm7.2}$} & \best{97.97$_{\pm2.2}$} & - & - & - & - & - & - \\
 & BrainMRI & 91.27$_{\pm3.2}$ & 98.01$_{\pm0.9}$ & 90.89$_{\pm2.1}$ & 97.95$_{\pm0.5}$ & \best{94.05$_{\pm2.1}$} & \best{98.78$_{\pm0.5}$} & 98.32$_{\pm0.3}$ & 80$_{\pm3.5}$ & 98.3$_{\pm0.2}$ & 79.33$_{\pm1.3}$ & \best{99.24$_{\pm0.1}$} & \best{87.52$_{\pm2.4}$} \\
 & Liver & 82.54$_{\pm5.8}$ & 82.95$_{\pm6.1}$ & 79.89$_{\pm0.6}$ & 78.86$_{\pm0.5}$ & \best{89.7$_{\pm1.3}$} & \best{89.34$_{\pm1.3}$} & 99.33$_{\pm0.6}$ & 55.23$_{\pm29.5}$ & 99.11$_{\pm0.7}$ & 54.3$_{\pm23.7}$ & \best{99.76$_{\pm0.2}$} & \best{69.57$_{\pm18.8}$} \\
 & RESC & \best{90.85$_{\pm4.7}$} & \best{88.63$_{\pm5.0}$} & 88.22$_{\pm4.0}$ & 84.91$_{\pm6.0}$ & 87.23$_{\pm3.9}$ & 85.86$_{\pm4.7}$ & 97.46$_{\pm0.8}$ & \best{74.65$_{\pm6.5}$} & 95.67$_{\pm0.8}$ & 64.27$_{\pm7.3}$ & \best{97.54$_{\pm0.8}$} & 73.33$_{\pm8.4}$ \\
\rowcolor{gray!20}  & AVG & 80.17$_{\pm8.5}$ & 88.03$_{\pm3.9}$ & 79.26$_{\pm7.0}$ & 86.49$_{\pm2.9}$ & \best{84.09$_{\pm5.2}$} & \best{89.56$_{\pm2.2}$} & 98.37$_{\pm0.5}$ & 69.96$_{\pm13.2}$ & 97.69$_{\pm0.6}$ & 65.97$_{\pm10.8}$ & \best{98.85$_{\pm0.4}$} & \best{76.8$_{\pm9.9}$} \\
\midrule
\multirow{7}{*}{\rotatebox{90}{2\,-\,shot}} & His & 75.83$_{\pm8.7}$ & 73.49$_{\pm7.8}$ & \best{77.01$_{\pm12.0}$} & \best{74.66$_{\pm11.1}$} & 73.98$_{\pm12.0}$ & 70.54$_{\pm12.5}$ & - & - & - & - & - & - \\
 & Chest & 67.81$_{\pm6.7}$ & 97.42$_{\pm1.0}$ & \best{72.22$_{\pm16.8}$} & 97.58$_{\pm2.2}$ & 69.33$_{\pm13.5}$ & \best{97.61$_{\pm1.1}$} & - & - & - & - & - & - \\
 & OCT17 & 94.53$_{\pm4.3}$ & 97.86$_{\pm1.9}$ & 96.43$_{\pm0.9}$ & 98.69$_{\pm0.4}$ & \best{98.43$_{\pm1.7}$} & \best{99.48$_{\pm0.6}$} & - & - & - & - & - & - \\
 & BrainMRI & 93.8$_{\pm1.0}$ & 98.56$_{\pm0.2}$ & 93.46$_{\pm1.1}$ & 98.54$_{\pm0.3}$ & \best{95.36$_{\pm1.0}$} & \best{99.03$_{\pm0.3}$} & 98.9$_{\pm0.1}$ & 82.42$_{\pm2.1}$ & 97.99$_{\pm0.2}$ & 81.55$_{\pm2.9}$ & \best{99.29$_{\pm0.2}$} & \best{87.36$_{\pm3.1}$} \\
 & Liver & 87.9$_{\pm5.0}$ & 88.14$_{\pm4.4}$ & 86.17$_{\pm2.4}$ & 86.67$_{\pm2.4}$ & \best{90.28$_{\pm3.8}$} & \best{90.51$_{\pm3.4}$} & 99.33$_{\pm0.9}$ & \best{64.82$_{\pm24.8}$} & 99.3$_{\pm0.3}$ & 60.48$_{\pm16.3}$ & \best{99.59$_{\pm0.3}$} & 62.89$_{\pm17.2}$ \\
 & RESC & \best{93.25$_{\pm0.4}$} & \best{91.92$_{\pm1.1}$} & 92.13$_{\pm1.3}$ & 90.46$_{\pm2.6}$ & 92.02$_{\pm1.9}$ & 91.11$_{\pm1.9}$ & 98.02$_{\pm0.6}$ & 81.75$_{\pm5.1}$ & 97.11$_{\pm0.5}$ & 78.74$_{\pm3.6}$ & \best{98.43$_{\pm0.4}$} & \best{82.05$_{\pm3.7}$} \\
\rowcolor{gray!20}  & AVG & 85.52$_{\pm4.3}$ & 91.23$_{\pm2.7}$ & 86.24$_{\pm5.7}$ & 91.1$_{\pm3.2}$ & \best{86.57$_{\pm5.7}$} & \best{91.38$_{\pm3.3}$} & 98.75$_{\pm0.5}$ & 76.33$_{\pm10.6}$ & 98.13$_{\pm0.4}$ & 73.59$_{\pm7.6}$ & \best{99.11$_{\pm0.3}$} & \best{77.43$_{\pm8.0}$} \\
\midrule
\multirow{7}{*}{\rotatebox{90}{4\,-\,shot}} & His & 74.38$_{\pm6.2}$ & 71.11$_{\pm8.1}$ & 77.61$_{\pm3.8}$ & 71.71$_{\pm4.3}$ & \best{77.62$_{\pm4.0}$} & \best{72.26$_{\pm4.1}$} & - & - & - & - & - & - \\
 & Chest & 75.92$_{\pm11.7}$ & 98.29$_{\pm1.0}$ & \best{83.57$_{\pm3.3}$} & \best{99$_{\pm0.2}$} & 78.74$_{\pm0.9}$ & 98.33$_{\pm0.0}$ & - & - & - & - & - & - \\
 & OCT17 & 99.11$_{\pm0.9}$ & 99.69$_{\pm0.3}$ & 98$_{\pm1.1}$ & 99.28$_{\pm0.4}$ & \best{99.9$_{\pm0.1}$} & \best{99.97$_{\pm0.0}$} & - & - & - & - & - & - \\
 & BrainMRI & 94.9$_{\pm0.5}$ & 98.81$_{\pm0.2}$ & 95.01$_{\pm0.2}$ & 98.93$_{\pm0.0}$ & \best{96.25$_{\pm0.6}$} & \best{99.22$_{\pm0.2}$} & 98.6$_{\pm0.5}$ & 84.17$_{\pm3.9}$ & 98.57$_{\pm0.5}$ & 84.07$_{\pm2.0}$ & \best{99.39$_{\pm0.1}$} & \best{88.74$_{\pm2.5}$} \\
 & Liver & 85.69$_{\pm6.3}$ & 82.99$_{\pm11.6}$ & 85.79$_{\pm2.0}$ & 85.72$_{\pm3.2}$ & \best{91.78$_{\pm2.1}$} & \best{92.4$_{\pm1.7}$} & 99.67$_{\pm0.2}$ & 76.43$_{\pm7.6}$ & 99.52$_{\pm0.2}$ & 77.35$_{\pm4.4}$ & \best{99.83$_{\pm0.2}$} & \best{82.43$_{\pm7.6}$} \\
 & RESC & \best{95.03$_{\pm1.2}$} & \best{93.62$_{\pm2.2}$} & 91.73$_{\pm5.0}$ & 90.24$_{\pm4.5}$ & 92.04$_{\pm1.9}$ & 91.43$_{\pm1.4}$ & \best{98.84$_{\pm0.5}$} & \best{87.65$_{\pm2.7}$} & 97.75$_{\pm1.1}$ & 83.02$_{\pm3.5}$ & 98.62$_{\pm0.3}$ & 83.89$_{\pm2.2}$ \\
\rowcolor{gray!20}  & AVG & 87.51$_{\pm4.5}$ & 90.75$_{\pm3.9}$ & 88.62$_{\pm2.6}$ & 90.81$_{\pm2.1}$ & \best{89.39$_{\pm1.6}$} & \best{92.27$_{\pm1.2}$} & 99.04$_{\pm0.4}$ & 82.75$_{\pm4.7}$ & 98.61$_{\pm0.6}$ & 81.48$_{\pm3.3}$ & \best{99.28$_{\pm0.2}$} & \best{85.02$_{\pm4.1}$} \\
\bottomrule
\end{tabular}

}
\end{table*}

\begin{table*}[t]
\centering
\caption{\textbf{Performance comparison under the one-for-many setting} for $k$ = 1, 2, and 4 shots, where a single model is jointly trained on all datasets and evaluated on each dataset individually. All methods (MVFA, MadCLIP, ProX-MAD) are evaluated with identical reference images per $k$-shot setting to ensure fair comparison. Image- and pixel-level results are averaged over three independently sampled few-shot reference sets (mean$_{\pm\text{std}}$). For the AVG row, both performance metrics and standard deviations are macro-averaged across datasets. $\Delta$ indicates the change in macro-averaged performance from the one-for-one setting to the one-for-many setting, where positive values denote performance gains. Best results are in \best{bold}.}
\label{tab:main_uni}
\small
\setlength{\tabcolsep}{3.5pt}
\renewcommand{\arraystretch}{1.15}
\setlength{\aboverulesep}{0pt}\setlength{\belowrulesep}{0pt}\setlength{\cmidrulesep}{0.5pt}
\resizebox{\linewidth}{!}{%

\begin{tabular}{ll|cccccc|cccccc}
\toprule
\multirow{3}{*}{$k$} & \multirow{3}{*}{Dataset}
 & \multicolumn{6}{c|}{Image-level} & \multicolumn{6}{c}{Pixel-level} \\
\cmidrule(lr){3-8}\cmidrule(lr){9-14}
 & & \multicolumn{2}{c}{MVFA~\cite{mvfa}} & \multicolumn{2}{c}{MadCLIP~\cite{madclip}} & \multicolumn{2}{c|}{\textbf{ProX-MAD}} & \multicolumn{2}{c}{MVFA~\cite{mvfa}} & \multicolumn{2}{c}{MadCLIP~\cite{madclip}} & \multicolumn{2}{c}{\textbf{ProX-MAD}} \\
\cmidrule(lr){3-4}\cmidrule(lr){5-6}\cmidrule(lr){7-8}
\cmidrule(lr){9-10}\cmidrule(lr){11-12}\cmidrule(lr){13-14}
 & & AUROC & AP & AUROC & AP & AUROC & AP & AUROC & AP & AUROC & AP & AUROC & AP \\
\midrule
\multirow{7}{*}{\rotatebox{90}{1\,-\,shot}} & His & 69.79$_{\pm6.8}$ & \best{70.65$_{\pm4.6}$} & \best{71.24$_{\pm8.5}$} & 70.43$_{\pm4.3}$ & 70.16$_{\pm6.7}$ & 65.09$_{\pm6.3}$ & - & - & - & - & - & - \\
 & Chest & 62.75$_{\pm14.7}$ & 96.83$_{\pm1.8}$ & 60.79$_{\pm16.2}$ & 96.73$_{\pm1.6}$ & \best{68.1$_{\pm10.8}$} & \best{97.75$_{\pm0.9}$} & - & - & - & - & - & - \\
 & OCT17 & 80.86$_{\pm21.0}$ & 92.36$_{\pm8.9}$ & 81.02$_{\pm21.4}$ & 92.44$_{\pm9.0}$ & \best{96.67$_{\pm1.2}$} & \best{98.88$_{\pm0.5}$} & - & - & - & - & - & - \\
 & BrainMRI & 89.73$_{\pm6.4}$ & 97.67$_{\pm1.7}$ & 90.13$_{\pm3.6}$ & 97.81$_{\pm0.9}$ & \best{93.38$_{\pm2.0}$} & \best{98.65$_{\pm0.4}$} & 98.24$_{\pm0.4}$ & 78.07$_{\pm6.0}$ & 96.36$_{\pm0.8}$ & 77.5$_{\pm2.1}$ & \best{99.15$_{\pm0.1}$} & \best{86.48$_{\pm1.9}$} \\
 & Liver & 80.4$_{\pm6.7}$ & 80.93$_{\pm6.8}$ & 77.66$_{\pm2.2}$ & 75.03$_{\pm2.1}$ & \best{90.37$_{\pm2.2}$} & \best{89.71$_{\pm2.3}$} & 99.4$_{\pm0.4}$ & 56.15$_{\pm23.0}$ & 95.69$_{\pm2.8}$ & 40.88$_{\pm29.4}$ & \best{99.8$_{\pm0.1}$} & \best{73.29$_{\pm14.8}$} \\
 & RESC & \best{90.16$_{\pm3.8}$} & \best{85.86$_{\pm4.2}$} & 87.33$_{\pm5.5}$ & 84.16$_{\pm6.6}$ & 86.8$_{\pm3.6}$ & 84.78$_{\pm4.9}$ & 96.58$_{\pm1.2}$ & 70.39$_{\pm10.4}$ & 93.11$_{\pm1.7}$ & 63.43$_{\pm8.3}$ & \best{97.44$_{\pm0.5}$} & \best{71.31$_{\pm7.7}$} \\
\rowcolor{gray!20}  & AVG & 78.95$_{\pm9.9}$ & 87.38$_{\pm4.7}$ & 78.03$_{\pm9.5}$ & 86.1$_{\pm4.1}$ & \best{84.25$_{\pm4.4}$} & \best{89.14$_{\pm2.5}$} & 98.07$_{\pm0.6}$ & 68.2$_{\pm13.1}$ & 95.05$_{\pm1.8}$ & 60.6$_{\pm13.3}$ & \best{98.8$_{\pm0.3}$} & \best{77.03$_{\pm8.1}$} \\
\rowcolor{gray!20}  & $\Delta$ & \red{-1.22} & \red{-0.64} & \red{-1.23} & \red{-0.39} & \blue{+0.16} & \red{-0.42} & \red{-0.29} & \red{-1.76} & \red{-2.64} & \red{-5.37} & \red{-0.05} & \blue{+0.22} \\
\midrule
\multirow{7}{*}{\rotatebox{90}{2\,-\,shot}} & His & 76.73$_{\pm3.8}$ & 71.09$_{\pm2.8}$ & \best{78.24$_{\pm10.2}$} & \best{76.23$_{\pm10.9}$} & 73.73$_{\pm6.8}$ & 71.01$_{\pm7.3}$ & - & - & - & - & - & - \\
 & Chest & 71.17$_{\pm12.2}$ & 97.66$_{\pm1.4}$ & \best{72.87$_{\pm13.8}$} & 97.72$_{\pm1.9}$ & 71.66$_{\pm4.3}$ & \best{97.83$_{\pm0.4}$} & - & - & - & - & - & - \\
 & OCT17 & 94.55$_{\pm1.6}$ & 98.09$_{\pm0.6}$ & 96.98$_{\pm1.1}$ & 98.99$_{\pm0.3}$ & \best{98.84$_{\pm0.6}$} & \best{99.63$_{\pm0.2}$} & - & - & - & - & - & - \\
 & BrainMRI & 92.16$_{\pm3.1}$ & 97.98$_{\pm1.2}$ & 92.71$_{\pm1.9}$ & 98.32$_{\pm0.5}$ & \best{95.15$_{\pm0.4}$} & \best{98.99$_{\pm0.1}$} & 98.58$_{\pm0.2}$ & 80$_{\pm1.9}$ & 94.82$_{\pm1.5}$ & 77.41$_{\pm5.2}$ & \best{99.28$_{\pm0.1}$} & \best{87.2$_{\pm1.5}$} \\
 & Liver & 86.24$_{\pm1.5}$ & 86.96$_{\pm1.6}$ & 84.06$_{\pm4.1}$ & 82.92$_{\pm4.9}$ & \best{89.45$_{\pm4.4}$} & \best{89.81$_{\pm4.1}$} & 99.29$_{\pm0.6}$ & 62.02$_{\pm14.3}$ & 96.3$_{\pm3.7}$ & 51.7$_{\pm27.1}$ & \best{99.64$_{\pm0.4}$} & \best{68.66$_{\pm22.7}$} \\
 & RESC & \best{92.21$_{\pm2.3}$} & 86.3$_{\pm7.4}$ & 91.93$_{\pm1.4}$ & 89.77$_{\pm3.1}$ & 91.39$_{\pm2.6}$ & \best{90.61$_{\pm2.5}$} & 98.01$_{\pm0.3}$ & \best{81.23$_{\pm2.6}$} & 96.68$_{\pm1.7}$ & 78$_{\pm4.1}$ & \best{98.16$_{\pm0.4}$} & 79.62$_{\pm4.8}$ \\
\rowcolor{gray!20}  & AVG & 85.51$_{\pm4.1}$ & 89.68$_{\pm2.5}$ & 86.13$_{\pm5.4}$ & 90.66$_{\pm3.6}$ & \best{86.7$_{\pm3.2}$} & \best{91.31$_{\pm2.4}$} & 98.63$_{\pm0.3}$ & 74.41$_{\pm6.2}$ & 95.94$_{\pm2.3}$ & 69.04$_{\pm12.1}$ & \best{99.03$_{\pm0.3}$} & \best{78.49$_{\pm9.7}$} \\
\rowcolor{gray!20}  & $\Delta$ & \red{-0.01} & \red{-1.55} & \red{-0.11} & \red{-0.44} & \blue{+0.13} & \red{-0.07} & \red{-0.12} & \red{-1.91} & \red{-2.20} & \red{-4.55} & \red{-0.08} & \blue{+1.06} \\
\midrule
\multirow{7}{*}{\rotatebox{90}{4\,-\,shot}} & His & 70.33$_{\pm8.6}$ & 67.65$_{\pm8.1}$ & 75.15$_{\pm5.1}$ & 70.39$_{\pm4.4}$ & \best{77.05$_{\pm6.6}$} & \best{72.31$_{\pm8.3}$} & - & - & - & - & - & - \\
 & Chest & 77.04$_{\pm8.3}$ & 98.4$_{\pm0.8}$ & \best{83.52$_{\pm3.1}$} & \best{99.0$_{\pm0.2}$} & 77.51$_{\pm5.1}$ & 98.23$_{\pm0.5}$ & - & - & - & - & - & - \\
 & OCT17 & 98.22$_{\pm2.4}$ & 99.42$_{\pm0.8}$ & 98.15$_{\pm1.2}$ & 99.39$_{\pm0.4}$ & \best{99.34$_{\pm0.6}$} & \best{99.79$_{\pm0.2}$} & - & - & - & - & - & - \\
 & BrainMRI & 94.51$_{\pm0.7}$ & 98.82$_{\pm0.2}$ & 94.82$_{\pm0.6}$ & 98.91$_{\pm0.1}$ & \best{95.98$_{\pm0.9}$} & \best{99.17$_{\pm0.2}$} & 98.71$_{\pm0.3}$ & 83.62$_{\pm3.6}$ & 95.89$_{\pm1.0}$ & 82.03$_{\pm3.3}$ & \best{99.36$_{\pm0.2}$} & \best{88.27$_{\pm2.5}$} \\
 & Liver & 88.05$_{\pm2.9}$ & 88.06$_{\pm3.1}$ & 85.06$_{\pm2.6}$ & 84.53$_{\pm4.1}$ & \best{92.56$_{\pm1.2}$} & \best{93.02$_{\pm1.2}$} & 99.72$_{\pm0.3}$ & 77.47$_{\pm10.5}$ & 98.8$_{\pm0.9}$ & 73.39$_{\pm7.6}$ & \best{99.88$_{\pm0.1}$} & \best{83.47$_{\pm6.2}$} \\
 & RESC & \best{93.63$_{\pm3.3}$} & \best{91.9$_{\pm4.5}$} & 92.18$_{\pm4.7}$ & 90.3$_{\pm4.3}$ & 91.75$_{\pm0.1}$ & 90.99$_{\pm0.8}$ & 98.26$_{\pm1.1}$ & \best{84.97$_{\pm5.2}$} & 97.13$_{\pm2.1}$ & 83.31$_{\pm3.9}$ & \best{98.46$_{\pm0.1}$} & 82.71$_{\pm1.1}$ \\
\rowcolor{gray!20}  & AVG & 86.96$_{\pm4.4}$ & 90.71$_{\pm2.9}$ & 88.15$_{\pm2.9}$ & 90.42$_{\pm2.3}$ & \best{89.03$_{\pm2.4}$} & \best{92.25$_{\pm1.9}$} & 98.9$_{\pm0.5}$ & 82.02$_{\pm6.4}$ & 97.27$_{\pm1.3}$ & 79.58$_{\pm4.9}$ & \best{99.23$_{\pm0.1}$} & \best{84.82$_{\pm3.3}$} \\
\rowcolor{gray!20}  & $\Delta$ & \red{-0.54} & \red{-0.04} & \red{-0.47} & \red{-0.40} & \red{-0.36} & \red{-0.02} & \red{-0.14} & \red{-0.73} & \red{-1.34} & \red{-1.90} & \red{-0.05} & \red{-0.20} \\
\bottomrule
\end{tabular}

}
\end{table*}

\subsection{Experimental Setups}

\textbf{Datasets and Evaluation Metrics.} We evaluate our method on six medical datasets from the BMAD benchmark~\cite{bmad}: Brain MRI~\cite{brainmri1,brainmri2,brainmri3}, Liver CT~\cite{liver,liver2}, OCT17~\cite{oct}, RESC~\cite{resc}, Chest X-ray~\cite{chest}, and Digital Histopathology~\cite{his}. We follow the evaluation protocol established by MVFA~\cite{mvfa} and MadCLIP~\cite{madclip}. Brain MRI, Liver, and RESC are used for both anomaly classification~(AC) and segmentation~(AS), while the others are evaluated solely for AC. We report both the area under the ROC curve (AUROC) and average precision (AP) at the image and pixel levels, denoted I-AUROC/P-AUROC and I-AP/P-AP. For datasets containing multiple lesion subtypes, we evaluate over all available lesion types rather than restricting the assessment to a single subtype.

\noindent \textbf{Implementation details.} We adopt a DINOv1 ViT-B/8~\cite{dino} as the visual encoder and resize all input images to $448\times448$. We extract multi-level patch features from the 3rd, 6th, 9th, and 12th layers. For each of these layers, we assign a layer-specific MPIA module (shared across states) and proxy tokens. Proxies are configured in two ways: one-for-one ($N=4$) and one-for-many ($N=24$). In the MPIA module, we apply masking with a ratio $\alpha$=0.9 to encourage robustness and diversity in proxy aggregation. We optimize the trainable parameters using AdamW with a learning rate of $2\times10^{-4}$. The training objective combines Dice, Focal, and BCE losses with weights $\lambda_1=1, \lambda_2=10$, and $\lambda_3=1$, respectively. All experiments are conducted on a single NVIDIA A6000 GPU (48GB).

\subsection{Comparison with State-of-the-art Methods}
\begin{figure*}[!t]%
  \centering
  \includegraphics[width=0.97\textwidth]{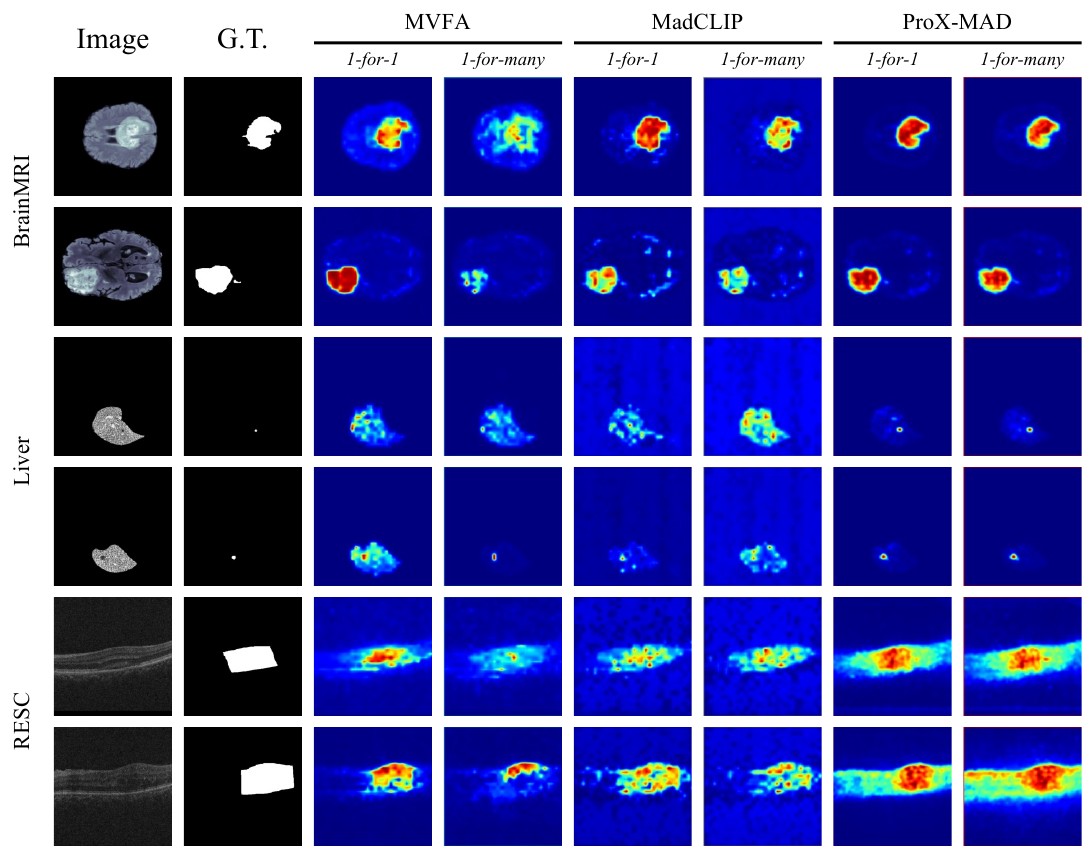}
  \caption{\textbf{Qualitative comparison of anomaly localization} against state-of-the-art methods under 4-shot settings.
}
  \label{fig:visual_cf}
\end{figure*}

\noindent\textbf{Comparison protocol.} We conduct comprehensive comparisons on six medical datasets under two evaluation protocols. The first is the one-for-one setting, where a separate model is trained and evaluated for each dataset. The second is the one-for-many setting, where a single model is trained on the aggregated few-shot data from all datasets and is then evaluated on each dataset without any additional dataset-specific fine-tuning. For fair comparisons, all methods are retrained using their official implementations under three fixed random seeds with identical few-shot samples, and we report the average performance.

\noindent\textbf{Main result.} \cref{tab:main_sep} and \cref{tab:main_uni} report the one-for-one and one-for-many settings, respectively. At both the image and pixel levels, ProX-MAD attains the best AVG performance at every shot budget in both settings: under the more demanding one-for-many setting it reaches $89.14/91.31/92.25\%$ I-AP and $77.03/78.49/84.82\%$ P-AP at $1/2/4$ shots, surpassing the strongest baseline at each budget (it likewise attains the best AVG AUROC in every case). The more revealing comparison is the transition from one-for-one to one-for-many, summarized in the ``$\Delta$'' rows. When a single model must serve all datasets, prior methods lose accuracy, and the loss is most severe at the pixel level: MadCLIP's AVG P-AP falls by $5.37/4.55/1.90$ and MVFA's by $1.76/1.91/0.73$ across the three budgets, because a shared decision boundary and a common reference bank cannot absorb the heterogeneous normal/abnormal appearances of different organs and modalities. ProX-MAD, in contrast, is nearly invariant to the setting: its AVG AP never falls by more than $0.42\%$, and it even improves at the pixel level for the $1$- and $2$-shot budgets ($+0.22$ and $+1.06$ P-AP). This stability arises because compact, state-specific proxies encode each dataset's normal and abnormal references without competing for a single boundary or a growing memory bank. The resulting minimal one-for-one/one-for-many gap provides evidence that a single ProX-MAD model can effectively serve multiple modalities seen during joint training.

\noindent\textbf{Qualitative comparison.} \cref{fig:visual_cf} contrasts anomaly localization in the one-for-one and one-for-many settings across datasets. While existing state-of-the-art methods localize lesions reasonably under the one-for-one setting, their predictions become diffuse or fragmented and drift away from the ground truth once moved to the one-for-many setting. ProX-MAD remains stable across both settings, producing sharp responses that stay well aligned with the ground truth. Moreover, even under the one-for-one setting where these methods are strongest, ProX-MAD delineates lesions more precisely. We attribute this to the combination of a DINO backbone that provides fine-grained spatial features and our proxy-based aggregation that turns them into state-specific prototypes.

\subsection{Ablation study}

\noindent Unless otherwise specified, all ablation experiments in this
subsection are conducted on the six datasets under the 4-shot one-for-many setting.

\begin{table}[t]
\centering
\small
\caption{\textbf{Ablation Study of Various Backbone Architectures.} Results are reported under the 1-/2-/4-shot one-for-many setting. $^\dagger$ indicates evaluation under the one-for-one setting.}
\label{tab:few_shot_results}
\resizebox{\textwidth}{!}{%
\begin{tabular}{l cccc cccc cccc}
\toprule
\multirow{2}{*}{Method} & 
\multicolumn{4}{c}{1-shot} &
\multicolumn{4}{c}{2-shot} &
\multicolumn{4}{c}{4-shot} \\
\cmidrule(lr){2-5} \cmidrule(lr){6-9} \cmidrule(lr){10-13}
 & I-AUROC & I-AP & P-AUROC & P-AP & I-AUROC & I-AP & P-AUROC & P-AP & I-AUROC & I-AP & P-AUROC & P-AP \\
\midrule
Previous SoTA$^\dagger$ & {80.17} & {88.03} & {98.37} & {69.96} 
            & {86.24} & {91.23} & {98.75} & {76.33} 
            & {88.62} & {90.81} & {99.04} & {82.75} \\
CLIP ViT-L/14-336 & 83.98 &	\best{90.26} &	{98.78} &	75.32
             & 86.66 &	\best{91.49} &	99.01 &	76.23
             & 88.81 &	92.05 &	99.2 &	80.77 \\
DINOv2 B/14-R & 82.21 &	89.15 &	98.47 &	72.57 
            & 86.39 &	90.89 &	\best{99.12} &	{78.32} 
            & 88.98 &	92.44 &	\textbf{99.25} &	82.65 \\
DINOv3 B/16 & 83.84 &	{89.67} &	98.36 &	71.72
            & 85.85 &	91.18 &	98.98 &	77.51
            & \best{89.03} &	\best{92.84} &	{99.16} &	82.77 \\
DINOv1 B/8 & \best{84.25} & {89.14} & \best{98.8} & \best{77.03} 
            & \best{86.7} & {91.31} & {99.03} & \best{78.49} 
            & \best{89.03} & {92.25} & {99.23} & \best{84.82} \\
\bottomrule
\end{tabular}%
}
\end{table}

\noindent\textbf{Effect of Backbone.} \cref{tab:few_shot_results} shows that our gains come from the design rather than the backbone: instantiated with the same CLIP ViT-L/14-336 encoder used by prior work, ProX-MAD trained in the one-for-many setting already matches or exceeds the previous one-for-one state of the art in I-AP at every shot budget (e.g., 90.26\% vs.\ 88.03\% at 1-shot, 92.05\% vs.\ 90.81\% at 4-shot), ruling out that the improvement merely stems from a stronger self-supervised encoder. Among backbones, DINOv1 ViT-B/8 achieves the best image-level AUROC and pixel-level AP across all shot budgets, despite its smaller parameter count, because its /8 tokenization produces a denser feature grid that preserves the fine-grained boundaries needed for small, irregular lesions.

\begin{figure*}[!t]%
  \centering
  \includegraphics[width=1\textwidth]{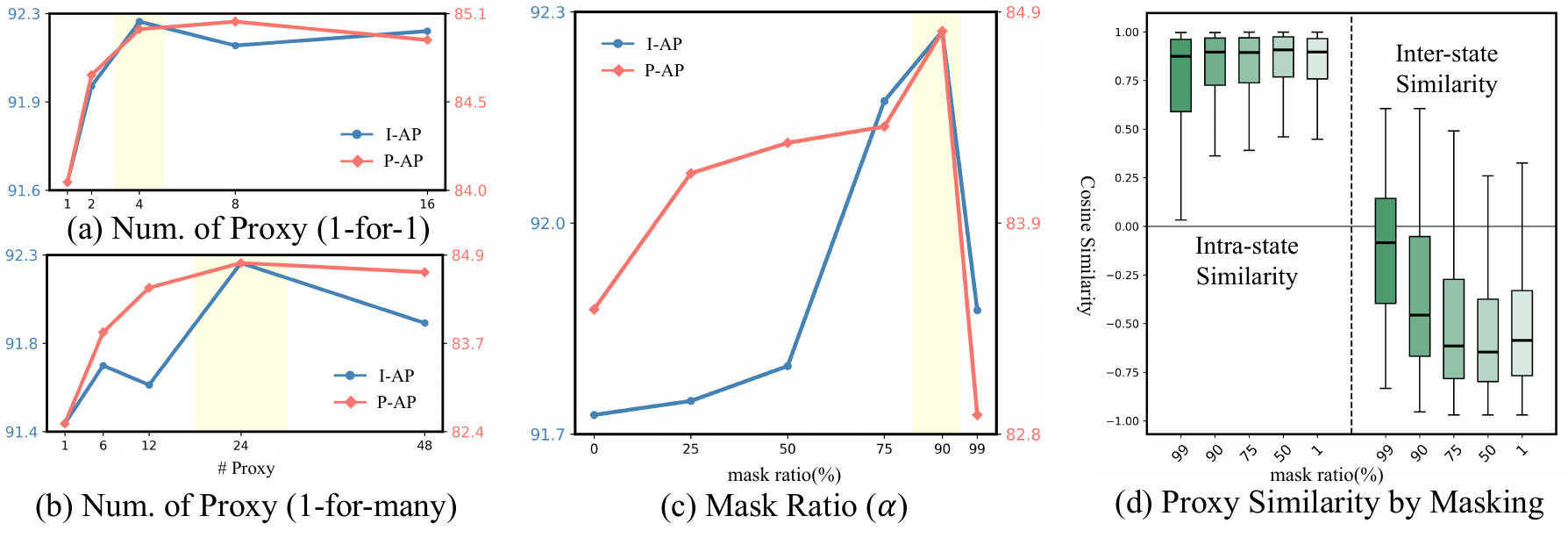}
  \caption{Ablation studies on key hyperparameters of the proposed ProX-MAD.}
  \label{fig:ablation}
\end{figure*}

\begin{table}[ht]
    \centering
    \begin{minipage}{0.39\textwidth}
        \centering
        \footnotesize
        \caption{Ablation Study of the Proposed Components. Frozen DINOv1 denotes a training-free baseline with a normal-reference memory bank.}
        \label{tab:ablation_components}
        \begin{tabular}{lccc}
            \hline
            Method & I-AP & P-AP \\ \hline
            Frozen DINOv1 & 82.18 & 42.77 \\
            {+ Proxy w/o MPIA} & 90.32 & 83.49 \\
            + Proxy w/ $F^n_{ref}$ & 91.91 & 83.97 \\
            \hspace{1.2mm} $\mathbf{\raisebox{0.2em}{$\llcorner$}}$ + Proxy w/ $F^a_{ref}$ & \best{92.25} & \best{84.82} \\
            \hline
        \end{tabular}
    \end{minipage}
    \hfill %
    \begin{minipage}{0.59\textwidth}
        \centering
        \footnotesize
        \caption{\textbf{Inference efficiency comparison.} All methods are evaluated using the same backbone under FP16 precision on a single NVIDIA A6000 GPU (48GB). $^\ddagger$ denotes memory-bank-based approaches.}
        \label{tab:ablation_backbone}
        \begin{tabular}{lcccc}
            \hline
            Method & \#train & latency(ms) & Mem.(mb) \\ \hline
            MVFA(1-shot)$^\ddagger$ & \best{12.58M} & 14.15 & 2522.3 \\
            MVFA(4-shot)$^\ddagger$ & \best{12.58M} & 15.01 & 2936.1 \\
            MadCLIP & {21.23M} & 13.79 & 2401.1 \\
            Ours & {29.59M} & \best{13.26} & \best{2126.9} \\
            \hline
        \end{tabular}
    \end{minipage}
\end{table}

\noindent\textbf{Hyper-parameter Ablation.} 
\cref{fig:ablation}(a) and (b) vary the number of proxies in the one-for-one and one-for-many settings, respectively: too few proxies fail to capture the intra-state diversity, and both I-AP and P-AP rise sharply from 1 to a few proxies, while gains saturate beyond 4 proxies in the one-for-one setting and 24 in the one-for-many setting, so we adopt these values. Notably, even when matched to a proxy count comparable to the one-for-one setting ($N=6$), ProX-MAD in the one-for-many setting still surpasses the previous one-for-one state of the art (90.81\% I-AP / 82.75\% P-AP). \cref{fig:ablation}(c) shows that increasing $\alpha$ from 0 to 0.9 yields a modest improvement in I-AP (+0.55) and a clearer gain in P-AP (+1.37), while overly aggressive masking at $\alpha=0.99$ degrades P-AP by discarding useful reference evidence. Fig. 4(d) further shows that masking reduces intra-state proxy similarity, consistent with mitigating proxy collapse. We therefore choose $\alpha{=}0.9$, which best balances accuracy and diversity.

\noindent\textbf{Core Design Ablation.} \cref{tab:ablation_components} isolates the effect of the proxy formulation and reference infusion, starting from a training-free \emph{Frozen DINOv1} baseline that scores queries against a memory bank of normal reference patches. It performs poorly (82.18\% I-AP / 42.77\% P-AP), consistent with limited normal coverage and weak normal--pathological separation in frozen features. Replacing the memory bank with learnable proxy tokens (+ Proxy w/o MPIA) lifts performance to 90.32\% / 83.49\%, showing that optimized proxies capture domain-specific structure without storing reference patches. Infusing normal references (+ Proxy w/ $F^n_{ref}$) raises I-AP to 91.91\%, while additionally infusing abnormal references (+ Proxy w/ $F^a_{ref}$) achieves the best performance (92.25\% / 84.82\%). With the reference set fixed, retraining across three seeds yields standard deviations of 0.51 I-AP and 0.64 P-AP in the 4-shot setting, both smaller than the corresponding gains from MPIA over plain proxies (+1.93 I-AP / +1.33 P-AP), supporting the benefit of reference infusion for learning more discriminative, state-specific representations.

\subsection{Analysis}

\noindent\textbf{Inference efficiency.} \cref{tab:ablation_backbone} compares inference cost under a common backbone on a single A6000 (fp16). ProX-MAD attains the lowest latency (13.26\,ms) and the smallest memory footprint (2126.9\,MB), outperforming both MVFA and MadCLIP. Although ProX-MAD trains more parameters (29.59\,M), these are learned once, are negligible beside the frozen VFM, and do not affect the inference profile. The key advantage is scaling with shots: MVFA relies on a memory bank of reference features, so its latency and memory grow as shots increase (14.15\,ms / 2522.3\,MB at 1-shot to 15.01\,ms / 2936.1\,MB at 4-shot), and this compounds when datasets are aggregated in the one-for-many setting. ProX-MAD instead summarizes the references into $N$ cached proxy tokens per state. Once $N$ is fixed for a trained model, its query-time cost does not increase with the number of reference shots, making it practical for unified multi-dataset deployment.

\begin{table}[t]
\centering
\small
\caption{\textbf{Cross dataset evaluation} under 1-shot setting. Results are obtained by training on dataset A and performing inference on dataset B ($A \rightarrow B$) without additional training. \best{Bold} indicates the best performance.}
\begin{tabular}{l cc cc cc cc}
\toprule
\multirow{3}{*}{Method} & 
\multicolumn{2}{c}{\makecell{Chest$\rightarrow$ \\CheXpert~\cite{Chexpert}}} &
\multicolumn{2}{c}{\makecell{RESC$\rightarrow$ \\OCTDL~\cite{octdl}}} &
\multicolumn{2}{c}{\makecell{OCT17$\rightarrow$ \\OCTDL~\cite{octdl}}} &
\multicolumn{2}{c}{AVG} \\
\cmidrule(lr){2-3} \cmidrule(lr){4-5} \cmidrule(lr){6-7} \cmidrule(lr){8-9}
 & I-AUROC & I-AP & I-AUROC & I-AP & I-AUROC & I-AP & I-AUROC & I-AP \\
\midrule
MVFA            & 62.9 & 64.21 & 78.91 & \best{95.22} & 83.74 & 96.53 & 75.18 & 85.32 \\
MadCLIP         & 65.94 & 66.1 & 73.88 & 93.09 & 79.3 & 94.73 & 73.04 & 84.64 \\
ProX-MAD~(Ours)  & \best{66.01} & \best{67.11} & \best{80.1} & {94.77} & \best{87.85} & \best{97.41} & \best{77.99} & \best{86.43} \\
\bottomrule
\end{tabular}
\label{tab:cross_data}
\end{table}
\noindent \textbf{Cross dataset evaluation.} \cref{tab:cross_data} evaluates transfer, training on dataset $A$ and testing on an unseen dataset $B$ ($A\rightarrow B$) with no further training. ProX-MAD attains the best I-AUROC on every transfer pair and the best average across both metrics (77.99\% I-AUROC / 86.43\% I-AP), ahead of MVFA (75.18\% / 85.32\%) and MadCLIP (73.04\% / 84.64\%). Because each source-target pair remains within the same modality while differing in scanner, acquisition, and lesion definition, these results support robustness under within-modality cross-dataset shifts.

\begin{figure*}[!t]
  \centering
  \includegraphics[width=1\textwidth]{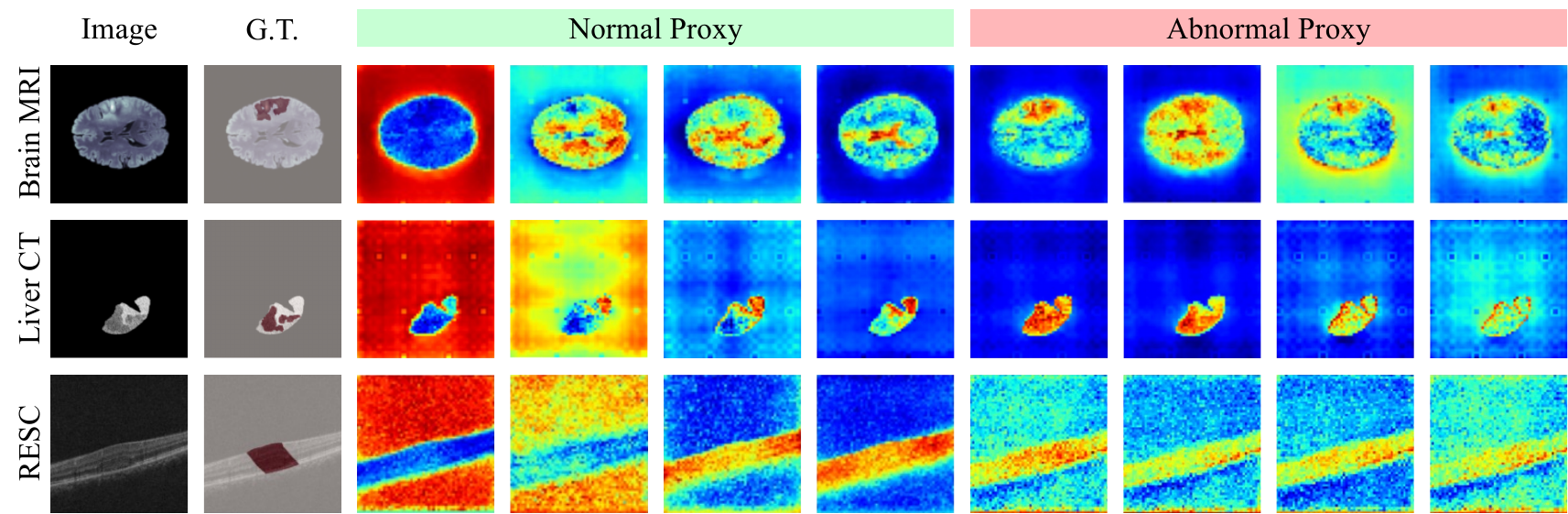}
  \caption{\textbf{Visualization of proxy activations.} Different proxies respond to different image patterns, demonstrating their ability to capture diverse characteristics associated with normality and abnormality.}
  \label{fig:proxy_vis}
  \vspace{-5mm}
\end{figure*}

\noindent\textbf{Proxy analysis.} \cref{fig:proxy_vis} shows, for each proxy, the query regions where its infused embedding attains high cosine similarity, revealing what each proxy attends to and captures. For the normal state, the proxies specialize along interpretable axes: some concentrate on the background and imaging field, while others attach to specific healthy anatomical regions, so together they tile the normal image rather than collapsing onto a single mode. This diversity follows directly from the proxy-wise masking in MPIA. For the abnormal state, the proxies distribute over complementary lesion characteristics, covering lesions of differing appearance, scale, and location rather than a single canonical anomaly, which is exactly what a unified model needs when lesion definitions vary across organs and modalities. Normal and abnormal proxies also remain well separated in embedding space, so the two sets act as contrasting anchors rather than redundant references. Overall, the proxies form a compact, diverse, and state-specific codebook that transfers across datasets. Beyond this, the $N$ normal and $N$ abnormal proxies also make our predictions more interpretable: because each proxy attends to a distinct, identifiable sub-pattern, a prediction can be traced back to the specific proxy that drove it, rather than relying on a single opaque similarity score.
\section{Conclusion}
\label{sec:conclusion}
\glsresetall

We presented \textbf{ProX-MAD}, a proxy-based framework for few-shot medical anomaly
detection under a one-for-many paradigm, where a single model is deployed across
multiple medical datasets without dataset-specific retraining. ProX-MAD summarizes
limited reference information into compact and learnable proxy tokens, and injects
state-matched reference cues via Masked Proxy Infusion Attention to obtain
representative and transferable prototypes for normal and abnormal states. By
computing similarity-based anomaly maps across selected intermediate layers and
optimizing with both pixel-level segmentation losses (when available) and an
image-level classification objective, our method achieves strong performance in
anomaly classification and anomaly segmentation. Extensive experiments show that ProX-MAD consistently achieves the best average performance under both one-for-one and one-for-many settings, and remains robust under cross-dataset evaluation, supporting its practicality for unified medical deployment. Because the number of proxy tokens can be flexibly adjusted, we expect ProX-MAD to scale to a substantially larger number of datasets within a single unified model.

\noindent\textbf{Limitation.}
Because ProX-MAD exploits rather than manufactures the separability of VFM features, it is less effective when normal and abnormal regions are not clearly separated in the feature space. This is most visible on RESC (analyzed in the supplementary), where ProX-MAD does not surpass the memory-bank baseline in image-level ranking, although it still attains the best pixel-level AUROC (\cref{tab:main_uni}).

\noindent\textbf{Future work.}
ProX-MAD still requires a few labeled references at test time; we plan to distill the learned proxies into text-conditioned prototypes using a DINO-style vision--language backbone~\cite{dinotxt}, enabling zero-shot deployment on new modalities without any reference image.

\section*{Acknowledgments} 
This work was supported by the National Research Foundation (NRF) grant funded by the Korea government (MSIT) [RS-2025-00562400] and [RS-2022-NR068754].

\clearpage

\bibliographystyle{splncs04}
\bibliography{reference}

\clearpage  %

\begin{center}
{\Large\bfseries Supplementary Material}\\[2pt]
{\large ProX-MAD: Aggregating Spatial Context into Representative Proxies\\[3pt]for Few-shot Medical Anomaly Detection}
\end{center}

\renewcommand{\thesection}{\Alph{section}}
\renewcommand{\thesubsection}{\thesection.\arabic{subsection}}
\renewcommand{\thesubsubsection}{\thesubsection.\arabic{subsubsection}}

\renewcommand{\thefigure}{S.\arabic{figure}}
\renewcommand{\thetable}{S.\arabic{table}}
\renewcommand{\theequation}{S.\arabic{equation}}

\setcounter{section}{0}
\setcounter{figure}{0}
\setcounter{table}{0}
\setcounter{equation}{0}

\makeatletter
\renewcommand{\theHsection}{supp.\Alph{section}}
\renewcommand{\theHsubsection}{\theHsection.\arabic{subsection}}
\renewcommand{\theHsubsubsection}{\theHsubsection.\arabic{subsubsection}}
\makeatother

\section{Dataset Details}

We evaluate ProX-MAD on six publicly available medical datasets provided by the BMAD benchmark~\cite{bmad}, spanning five anatomical regions and multiple imaging modalities (MRI, CT, OCT, X-ray, and histopathology). This collection is intentionally heterogeneous in acquisition physics, anatomical structure, and lesion definition, which makes it a stringent testbed for the unified one-for-many setting studied in this work. \cref{tab_supp:dataset_summary} reports, for each dataset, its anatomical category, test-set composition, and availability of pixel-annotation. Three datasets (BrainMRI~\cite{brainmri1,brainmri2,brainmri3}, Liver~\cite{liver,liver2} and RESC~\cite{resc}) provide ground-truth masks at the pixel-level and are therefore evaluated for both anomaly classification (AC) and anomaly segmentation (AS); the remaining three (His~\cite{his}, Chest~\cite{chest}, and OCT17~\cite{oct}) provide only image-level labels and are evaluated for AC alone. For all six datasets, we adopt the standard BMAD train/test split and use only its designated test set for evaluation, while the few-shot reference pool ($k$ normal and $k$ abnormal samples, $k\in\{1,2,4\}$) is drawn as described in \cref{sec:method}.

\begin{table}[]
\centering
\caption{\textbf{Summary of the six BMAD datasets used in our experiments.} For each dataset we report its anatomical category, the number of normal and abnormal images in the test set, and whether pixel-level annotations are available (\ding{51}) or not (\ding{55}). Datasets with pixel annotations are evaluated for both AC and AS; the others for AC only.}
\label{tab_supp:dataset_summary}
\begin{tabular}{lcccc}
\toprule
Dataset & Category & Test Normal & Test Abnormal & Pixel Annotation \\
\midrule
His       & Lymph Node (Breast)      & 1,003 & 994& \ding{55} \\
Chest     & Lung                   & 781      & 16,413& \ding{55}  \\
OCT17     & Retina                   & 242 & 726& \ding{55}  \\
BrainMRI  & Brain                    & 640     & 3,075& \ding{51} \\
Liver     & Liver                   & 833     & 660& \ding{51} \\
RESC      & Retina                  & 1,041 & 764& \ding{51} \\
\bottomrule
\end{tabular}
\end{table}

\section{Per-Dataset Backbone Ablation}

\providecommand{\best}[1]{\textbf{#1}}
\providecommand{\ph}{\textcolor{gray}{--}}   %
\begin{table*}[t]
\centering
\caption{\textbf{Performance comparison across various backbones under the one-for-many setting} for $k$ = 1, 2, and 4 shots. All backbone variants of ProX-MAD are evaluated with identical few-shot reference images per $k$-shot setting to ensure fair comparison. Image- and pixel-level (AUROC, AP) results are averaged over 3 runs (mean$_{\pm\text{std}}$); For the AVG row, both performance metrics and standard deviations are macro-averaged across datasets. Best results are in \best{bold}.}
\label{tab_supp:backbone}
\small
\setlength{\tabcolsep}{3.5pt}
\renewcommand{\arraystretch}{1.15}
\setlength{\aboverulesep}{0pt}\setlength{\belowrulesep}{0pt}\setlength{\cmidrulesep}{0.5pt}
\resizebox{\linewidth}{!}{%

\begin{tabular}{ll|cccccc|cccccc}
\toprule
\multirow{3}{*}{$k$} & \multirow{3}{*}{Dataset}
 & \multicolumn{6}{c|}{Image-level} & \multicolumn{6}{c}{Pixel-level} \\
\cmidrule(lr){3-8}\cmidrule(lr){9-14}
 & & \multicolumn{2}{c}{CLIP ViT-L/14-336} & \multicolumn{2}{c}{DINOv2 B/14-R} & \multicolumn{2}{c|}{DINOv3 B/16} & \multicolumn{2}{c}{CLIP ViT-L/14-336} & \multicolumn{2}{c}{DINOv2 B/14-R} & \multicolumn{2}{c}{DINOv3 B/16} \\
\cmidrule(lr){3-4}\cmidrule(lr){5-6}\cmidrule(lr){7-8}
\cmidrule(lr){9-10}\cmidrule(lr){11-12}\cmidrule(lr){13-14}
 & & AUROC & AP & AUROC & AP & AUROC & AP & AUROC & AP & AUROC & AP & AUROC & AP \\
\midrule
\multirow{7}{*}{\rotatebox{90}{1\,-\,shot}} & His & 69.11$_{\pm4.4}$ & 66.30$_{\pm3.4}$ & 67.69$_{\pm3.6}$ & 65.74$_{\pm2.1}$ & \best{69.37$_{\pm1.4}$} & \best{66.58$_{\pm1.1}$} & - & - & - & - & - & - \\
 & Chest & 64.01$_{\pm12.9}$ & 96.79$_{\pm1.4}$ & 64.34$_{\pm8.0}$ & 97.31$_{\pm0.7}$ & \best{67.77$_{\pm9.4}$} & \best{97.72$_{\pm0.8}$} & - & - & - & - & - & - \\
 & OCT17 & 95.94$_{\pm2.7}$ & 98.70$_{\pm0.9}$ & 95.79$_{\pm6.5}$ & 98.62$_{\pm2.1}$ & \best{97.55$_{\pm2.6}$} & \best{99.25$_{\pm0.8}$} & - & - & - & - & - & - \\
 & BrainMRI & \best{93.24$_{\pm1.4}$} & \best{98.62$_{\pm0.3}$} & 87.56$_{\pm0.7}$ & 97.19$_{\pm0.1}$ & 90.44$_{\pm0.7}$ & 97.91$_{\pm0.2}$ & \best{98.98$_{\pm0.2}$} & \best{84.26$_{\pm2.2}$} & 98.46$_{\pm0.3}$ & 77.30$_{\pm3.0}$ & 98.35$_{\pm0.2}$ & 76.90$_{\pm1.9}$ \\
 & Liver & 89.83$_{\pm2.4}$ & 90.10$_{\pm2.0}$ & \best{89.87$_{\pm4.2}$} & \best{90.19$_{\pm3.6}$} & 87.49$_{\pm4.1}$ & 87.97$_{\pm3.6}$ & 99.55$_{\pm0.1}$ & 64.88$_{\pm20.9}$ & \best{99.76$_{\pm0.0}$} & \best{71.12$_{\pm9.8}$} & 99.72$_{\pm0.1}$ & 67.30$_{\pm15.8}$ \\
 & RESC & \best{91.73$_{\pm2.4}$} & \best{91.04$_{\pm2.3}$} & 88.03$_{\pm1.6}$ & 85.87$_{\pm1.1}$ & 90.41$_{\pm1.2}$ & 88.57$_{\pm1.5}$ & \best{97.81$_{\pm0.3}$} & \best{76.83$_{\pm1.5}$} & 97.19$_{\pm0.2}$ & 69.29$_{\pm4.4}$ & 97.02$_{\pm0.1}$ & 70.96$_{\pm0.8}$ \\
\rowcolor{gray!20}  & AVG & \best{83.98$_{\pm4.4}$} & \best{90.26$_{\pm1.7}$} & 82.21$_{\pm4.1}$ & 89.15$_{\pm1.6}$ & 83.84$_{\pm3.2}$ & 89.67$_{\pm1.3}$ & \best{98.78$_{\pm0.2}$} & \best{75.32$_{\pm8.2}$} & 98.47$_{\pm0.2}$ & 72.57$_{\pm5.7}$ & 98.36$_{\pm0.2}$ & 71.72$_{\pm6.2}$ \\
\midrule
\multirow{7}{*}{\rotatebox{90}{2\,-\,shot}} & His & \best{71.85$_{\pm3.1}$} & \best{68.56$_{\pm3.1}$} & 64.24$_{\pm5.0}$ & 61.74$_{\pm4.3}$ & 69.51$_{\pm3.1}$ & 66.55$_{\pm3.3}$ & - & - & - & - & - & - \\
 & Chest & \best{70.72$_{\pm8.6}$} & \best{97.76$_{\pm0.8}$} & 69.77$_{\pm5.7}$ & 97.59$_{\pm0.5}$ & 65.94$_{\pm9.8}$ & 97.21$_{\pm1.1}$ & - & - & - & - & - & - \\
 & OCT17 & 98.90$_{\pm0.7}$ & 99.73$_{\pm0.3}$ & \best{99.82$_{\pm0.2}$} & \best{99.95$_{\pm0.1}$} & 99.59$_{\pm0.7}$ & 99.87$_{\pm0.2}$ & - & - & - & - & - & - \\
 & BrainMRI & 94.85$_{\pm1.4}$ & 98.93$_{\pm0.4}$ & \best{96.43$_{\pm0.2}$} & \best{99.25$_{\pm0.1}$} & 95.77$_{\pm0.7}$ & 99.09$_{\pm0.2}$ & 99.23$_{\pm0.1}$ & \best{86.04$_{\pm2.5}$} & \best{99.28$_{\pm0.0}$} & 85.82$_{\pm0.2}$ & 99.12$_{\pm0.0}$ & 85.62$_{\pm1.2}$ \\
 & Liver & 89.58$_{\pm2.7}$ & 91.14$_{\pm2.4}$ & \best{93.11$_{\pm1.3}$} & \best{93.19$_{\pm0.8}$} & 89.21$_{\pm1.9}$ & 90.07$_{\pm1.5}$ & 99.55$_{\pm0.3}$ & 62.44$_{\pm16.7}$ & \best{99.76$_{\pm0.1}$} & \best{68.22$_{\pm7.4}$} & 99.70$_{\pm0.1}$ & 66.74$_{\pm8.8}$ \\
 & RESC & 94.03$_{\pm0.2}$ & 92.83$_{\pm0.9}$ & 94.99$_{\pm1.3}$ & 93.63$_{\pm2.1}$ & \best{95.05$_{\pm1.2}$} & \best{94.27$_{\pm1.8}$} & 98.26$_{\pm0.4}$ & 80.20$_{\pm4.2}$ & \best{98.31$_{\pm0.5}$} & \best{80.92$_{\pm3.8}$} & 98.11$_{\pm0.8}$ & 80.16$_{\pm6.3}$ \\
\rowcolor{gray!20}  & AVG & \best{86.66$_{\pm2.8}$} & \best{91.49$_{\pm1.3}$} & 86.39$_{\pm2.3}$ & 90.89$_{\pm1.3}$ & 85.85$_{\pm2.9}$ & 91.18$_{\pm1.3}$ & 99.01$_{\pm0.3}$ & 76.23$_{\pm7.8}$ & \best{99.12$_{\pm0.2}$} & \best{78.32$_{\pm3.8}$} & 98.98$_{\pm0.3}$ & 77.51$_{\pm5.4}$ \\
\midrule
\multirow{7}{*}{\rotatebox{90}{4\,-\,shot}} & His & 74.13$_{\pm6.1}$ & 69.69$_{\pm3.8}$ & 74.81$_{\pm3.0}$ & 69.60$_{\pm2.0}$ & \best{75.80$_{\pm4.9}$} & \best{71.38$_{\pm3.7}$} & - & - & - & - & - & - \\
 & Chest & \best{77.59$_{\pm3.5}$} & \best{98.30$_{\pm0.2}$} & 74.52$_{\pm0.8}$ & 97.99$_{\pm0.1}$ & 74.45$_{\pm1.2}$ & 98.03$_{\pm0.2}$ & - & - & - & - & - & - \\
 & OCT17 & 99.75$_{\pm0.2}$ & 99.92$_{\pm0.1}$ & 99.64$_{\pm0.3}$ & 99.89$_{\pm0.1}$ & \best{99.88$_{\pm0.1}$} & \best{99.96$_{\pm0.0}$} & - & - & - & - & - & - \\
 & BrainMRI & 96.27$_{\pm0.4}$ & 99.22$_{\pm0.1}$ & \best{96.88$_{\pm0.2}$} & 99.36$_{\pm0.0}$ & 96.75$_{\pm0.1}$ & \best{99.37$_{\pm0.1}$} & 99.27$_{\pm0.1}$ & 85.82$_{\pm2.6}$ & \best{99.40$_{\pm0.0}$} & \best{88.35$_{\pm0.7}$} & 99.15$_{\pm0.1}$ & 87.42$_{\pm2.0}$ \\
 & Liver & 90.35$_{\pm1.6}$ & 91.45$_{\pm1.4}$ & \best{93.25$_{\pm1.8}$} & \best{94.01$_{\pm1.3}$} & 92.13$_{\pm0.7}$ & 93.38$_{\pm1.2}$ & 99.75$_{\pm0.1}$ & 73.91$_{\pm2.0}$ & \best{99.78$_{\pm0.1}$} & 76.37$_{\pm4.4}$ & 99.76$_{\pm0.1}$ & \best{76.78$_{\pm0.9}$} \\
 & RESC & 94.79$_{\pm2.2}$ & 93.70$_{\pm2.4}$ & 94.76$_{\pm1.4}$ & 93.81$_{\pm2.6}$ & \best{95.19$_{\pm0.6}$} & \best{94.92$_{\pm1.0}$} & \best{98.58$_{\pm0.4}$} & 82.59$_{\pm1.9}$ & 98.56$_{\pm0.3}$ & 83.23$_{\pm3.3}$ & 98.57$_{\pm0.1}$ & \best{84.10$_{\pm3.2}$} \\
\rowcolor{gray!20}  & AVG & 88.81$_{\pm2.3}$ & 92.05$_{\pm1.3}$ & 88.98$_{\pm1.2}$ & 92.44$_{\pm1.0}$ & \best{89.03$_{\pm1.2}$} & \best{92.84$_{\pm1.0}$} & 99.20$_{\pm0.2}$ & 80.77$_{\pm2.2}$ & \best{99.25$_{\pm0.1}$} & 82.65$_{\pm2.8}$ & 99.16$_{\pm0.1}$ & \best{82.77$_{\pm2.0}$} \\
\bottomrule
\end{tabular}

}
\end{table*}

\cref{tab:few_shot_results} in the main paper summarizes the effect of the
visual backbone using dataset-averaged scores. Here we provide the full
per-dataset breakdown behind that summary. \cref{tab_supp:backbone} reports,
for each of the six datasets, the image- and pixel-level AUROC and AP obtained
when ProX-MAD is instantiated with three alternative frozen encoders
(CLIP ViT-L/14-336~\cite{clip}, DINOv2 ViT-B/14-R~\cite{vit_reg}, and DINOv3 ViT-B/16~\cite{dinov3}). All three variants are trained under
the one-for-many setting with identical few-shot references. The per-dataset
results for our default DINOv1 ViT-B/8~\cite{dino} backbone are already listed alongside the baselines in the main results table (\cref{tab:main_uni}); we therefore omit them here to avoid duplication and focus this table on the alternative encoders.

\noindent\textbf{No single alternative backbone dominates.}
Across shot budgets, the best alternative encoder changes: CLIP attains the
highest average image-level AP at 1-shot ($90.26\%$), whereas DINOv3 becomes the
strongest at 4-shot ($92.84\%$ I-AP, $82.77\%$ P-AP). This instability across budgets contrasts with our default DINOv1 ViT-B/8, which the main paper shows to achieve the best image-level AUROC and pixel-level AP across all shot budgets. The gap is consistent with our motivation
(\cref{sec:intro}): the finer /8 tokenization of DINOv1 yields a denser feature
grid that better preserves the boundaries of small,  irregular lesions; the per-dataset instability of the alternative encoders below is consistent with this view. 

\noindent\textbf{Per-dataset heterogeneity mirrors the one-for-many challenge.}
No encoder is uniformly best across datasets, reflecting the domain heterogeneity
that a unified model must absorb. At 1-shot, CLIP leads on BrainMRI and RESC at
both the image and pixel levels, while DINOv2 is strongest on Liver; at 4-shot,
DINOv3 takes the lead on the retinal and histopathology datasets. This scattering
of per-dataset winners is precisely why we do not rely on any single alternative
encoder and instead adopt the backbone that is most consistent across domains.

\section{Feature Separability Analysis via PCA}
\begin{figure}[h!]
    \centering
    \includegraphics[width=0.85\linewidth]{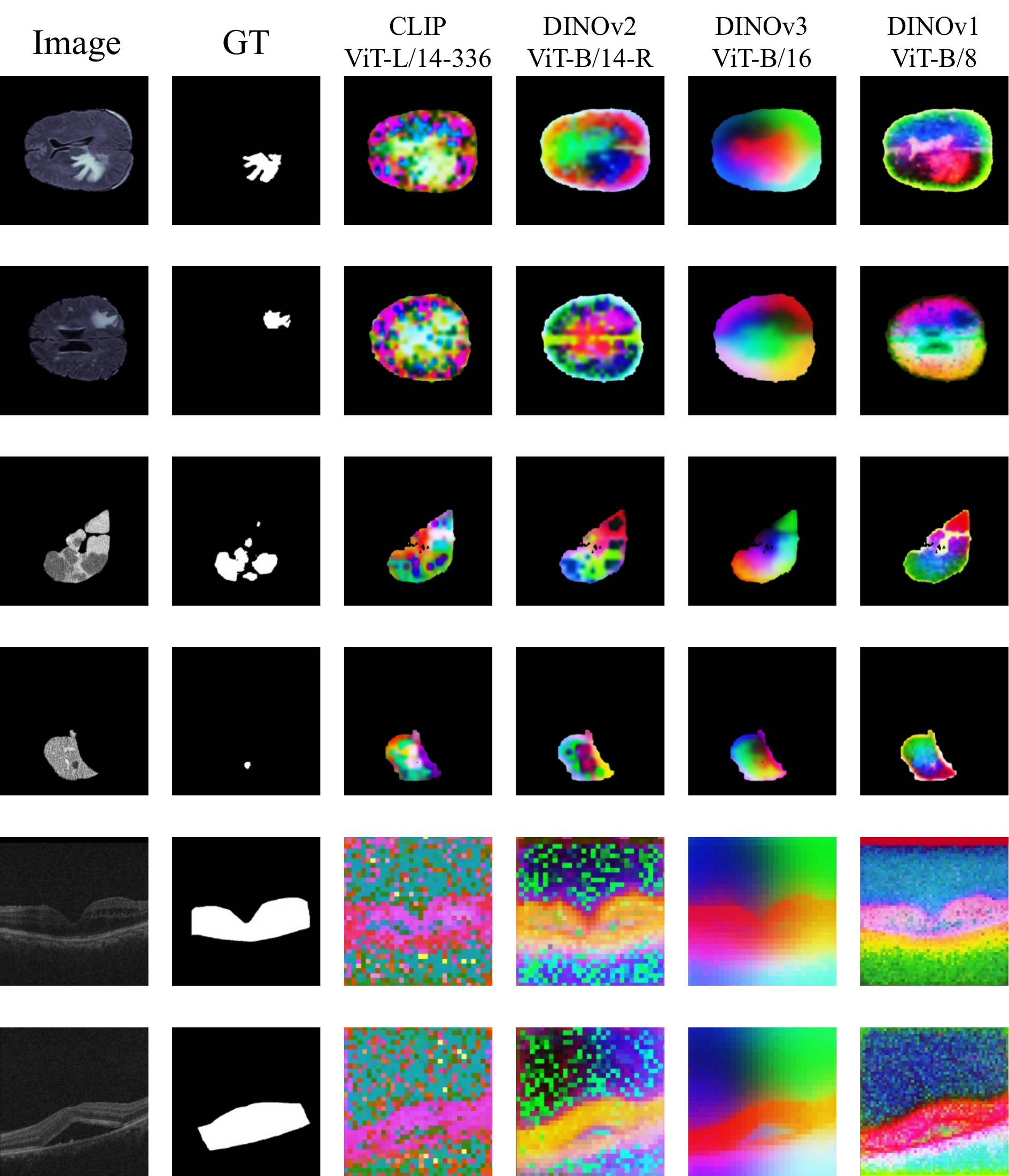}
    \caption{\textbf{PCA visualization of patch-level features from frozen vision foundation models.} For each sample (rows: brain MRI, liver CT, and RESC), foreground patch features from CLIP ViT-L/14-336, DINOv2 ViT-B/14-R, DINOv3 ViT-B/16, and DINOv1 ViT-B/8 are projected onto their first three principal components and shown as RGB. Coherent color regions indicate features that separate tissue types, while noisy coloring indicates entangled features. On brain and liver, lesions form a distinct cluster separable from healthy tissue; on RESC, features are organized along retinal layers and the edematous region is not clearly separated, illustrating the limited intrinsic separability discussed in \cref{sec:conclusion}.}
    \label{fig_supp:detail_pca_anly}
\end{figure}

To complement the teaser analysis in the main paper (\cref{fig:teaser}(d)) and to
substantiate the limitation discussed in \cref{sec:conclusion}, we visualize the
patch-level features of several frozen vision foundation models via PCA.
For each image, we extract dense patch features from the final layer of each
backbone, restrict them to the foreground (the imaged anatomy), and project them
onto their first three principal components, which are mapped to the RGB channels.
Under this mapping, patches that share a color occupy a similar region of feature
space, so a backbone that separates tissue types well produces spatially coherent,
well-delineated color regions, whereas a backbone whose features are entangled
produces noisy or diffuse coloring.

\cref{fig_supp:detail_pca_anly} shows this visualization across brain MRI, liver CT, and
retinal OCT (RESC) samples. Two observations stand out. First, consistent with
our motivation, the CLIP features are the least spatially structured: normal and
lesion regions are not cleanly separated and the projection appears noisy,
especially on the retinal scans. DINO-family backbones yield markedly more
coherent structure, and the fine /8 tokenization of DINOv1 produces the sharpest,
most localized boundaries, which is advantageous for the small, irregular lesions
common in medical imaging.

Second, central to our claimed limitation, the RESC (retinal OCT) rows differ markedly from the brain and liver rows. On brain MRI and liver CT, the lesion
region emerges as a distinct color cluster that stands apart from the surrounding
healthy tissue, i.e., normal and abnormal patches are separable in feature space.
On RESC, however, even the strongest backbones organize the features primarily
along the layered retinal structure rather than around the edema itself: feature
values vary smoothly across retinal layers, and the edematous region does not form
a color cluster that is clearly disjoint from the normal retina. Because ProX-MAD
exploits, rather than manufactures, the separability already present in the
backbone features, this weak intrinsic separation on RESC explains why our method
does not surpass the memory-bank baseline in image-level ranking on this dataset,
even though it still attains the best pixel-level AUROC (\cref{tab:main_uni}).

\section{Additional Qualitative Results}

We provide extended qualitative comparisons of ProX-MAD across the three datasets
with pixel-level annotations (BrainMRI, Liver, and RESC) under the $1$-, $2$-, and
$4$-shot settings. In every figure, each column corresponds to one test sample and
the four rows show, from top to bottom, the input image, the ground-truth lesion
mask (GT), the anomaly map produced by our dataset-specific \emph{one-for-one}
model, and the anomaly map produced by our single unified \emph{one-for-many}
model. Warmer colors denote higher predicted anomaly scores. These figures isolate
the effect of moving from dataset-specific to unified training: because both rows
share the same ProX-MAD design and identical references (differing only in the number of proxies), the similarity between the
one-for-one and one-for-many maps directly reflects how little our method degrades
when a single model must serve all datasets at once. Across datasets and shot
budgets, the one-for-many maps remain closely aligned with both the one-for-one
maps and the ground truth, localizing lesions of varying size, shape, and count
with substantially less diffusion or fragmentation than typically induced by unified training.

\begin{figure}
    \centering
    \includegraphics[width=1.0\linewidth]{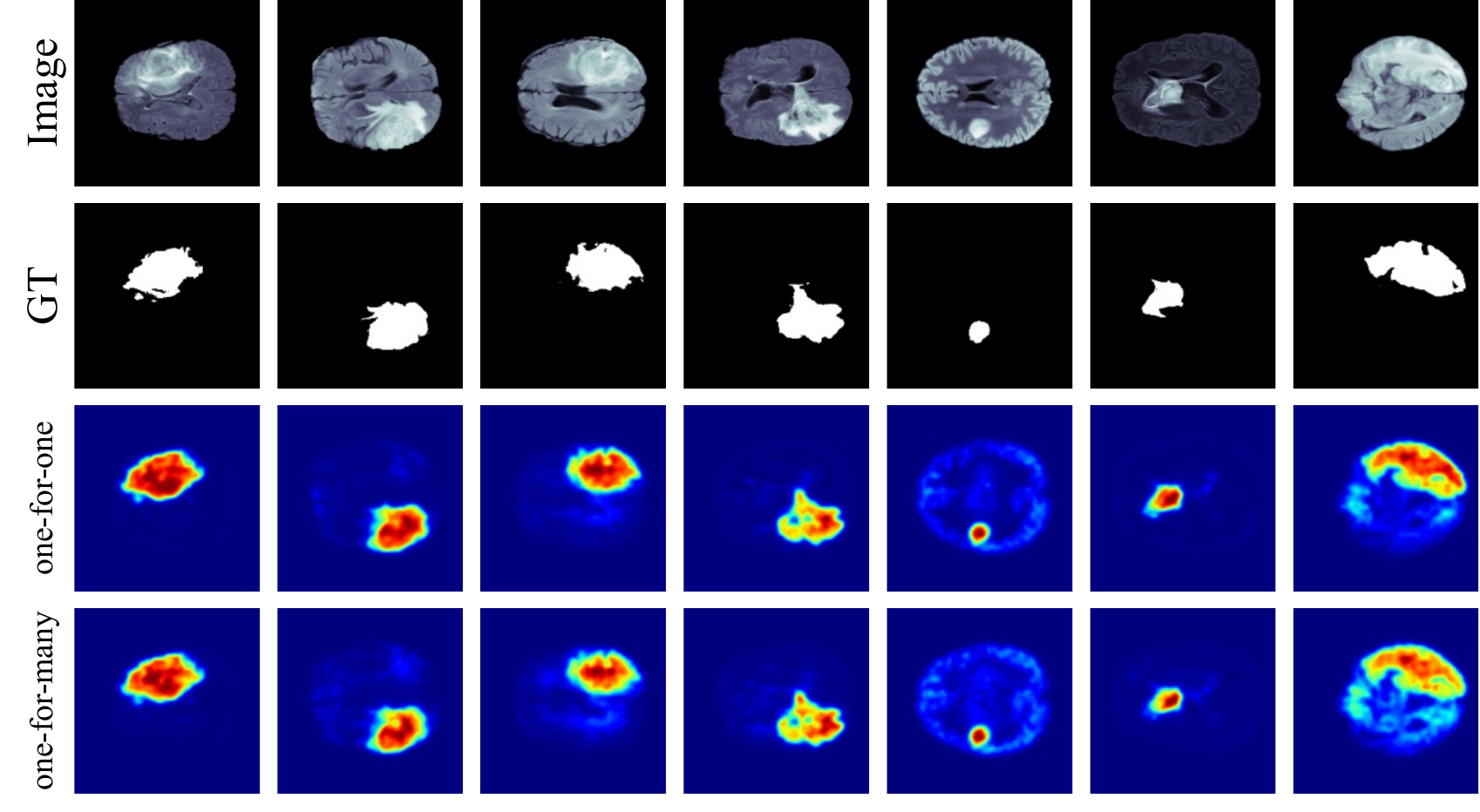}
    \caption{Qualitative anomaly localization on \textbf{BrainMRI} under the
    \textbf{1-shot} setting. Rows: input, GT, one-for-one, one-for-many.}
    \label{fig_supp:brain_1}
\end{figure}
\begin{figure}
    \centering
    \includegraphics[width=1.0\linewidth]{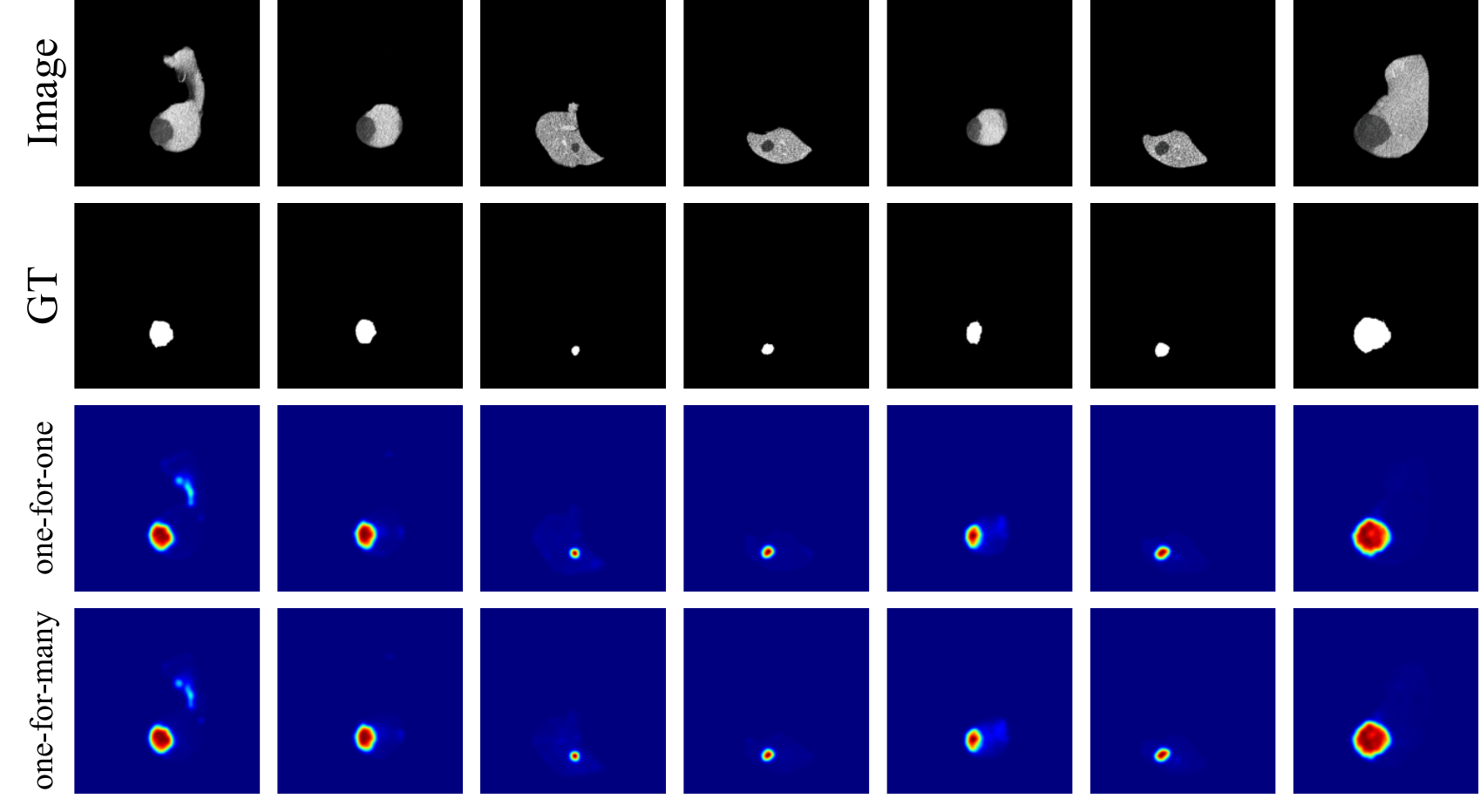}
    \caption{Qualitative anomaly localization on \textbf{Liver} under the
    \textbf{1-shot} setting. Rows: input, GT, one-for-one, one-for-many.}
    \label{fig_supp:liver_1}
\end{figure}
\begin{figure}
    \centering
    \includegraphics[width=1.0\linewidth]{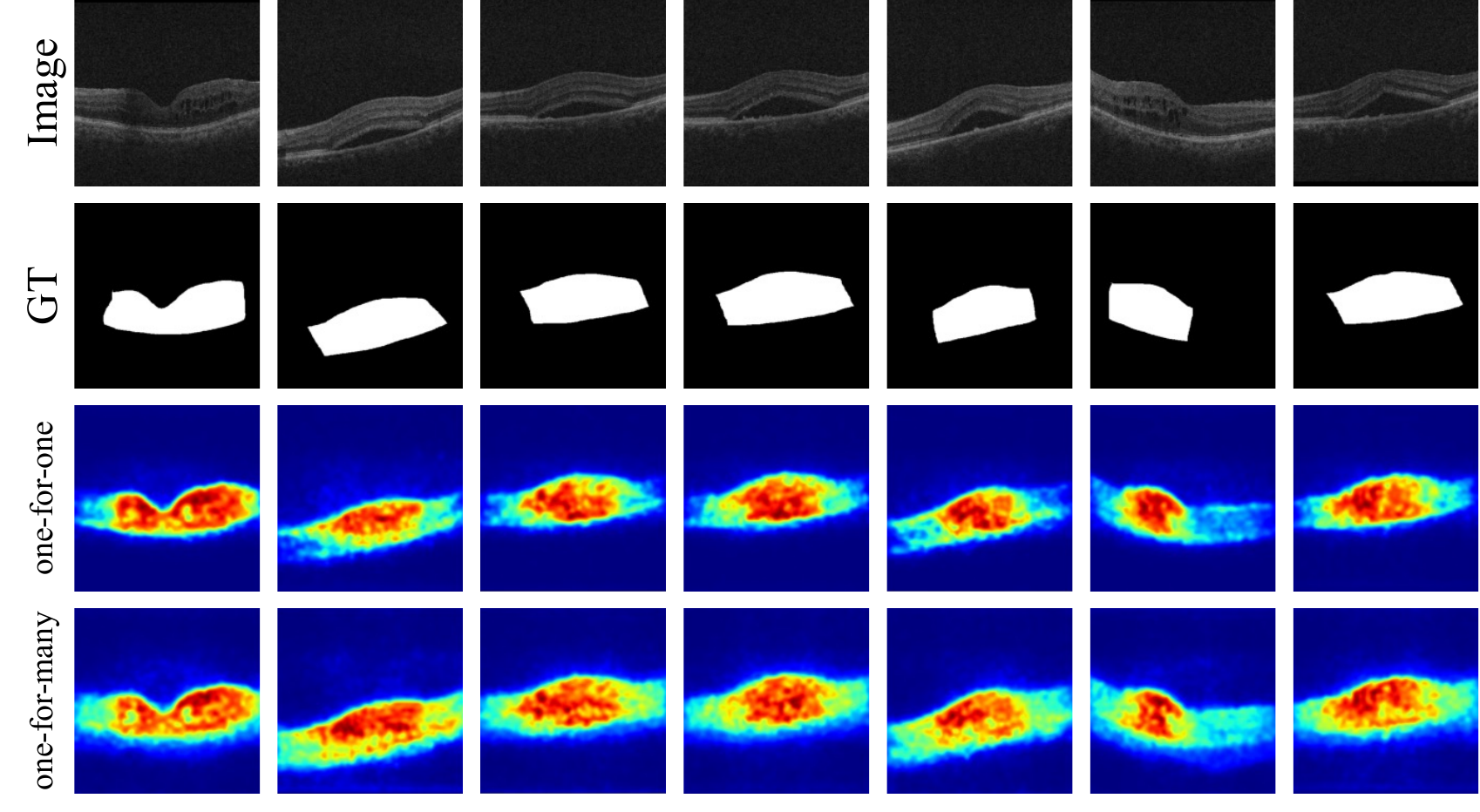}
    \caption{Qualitative anomaly localization on \textbf{RESC} under the
    \textbf{1-shot} setting. Rows: input, GT, one-for-one, one-for-many.}
    \label{fig_supp:resc_1}
\end{figure}

\begin{figure}
    \centering
    \includegraphics[width=1.0\linewidth]{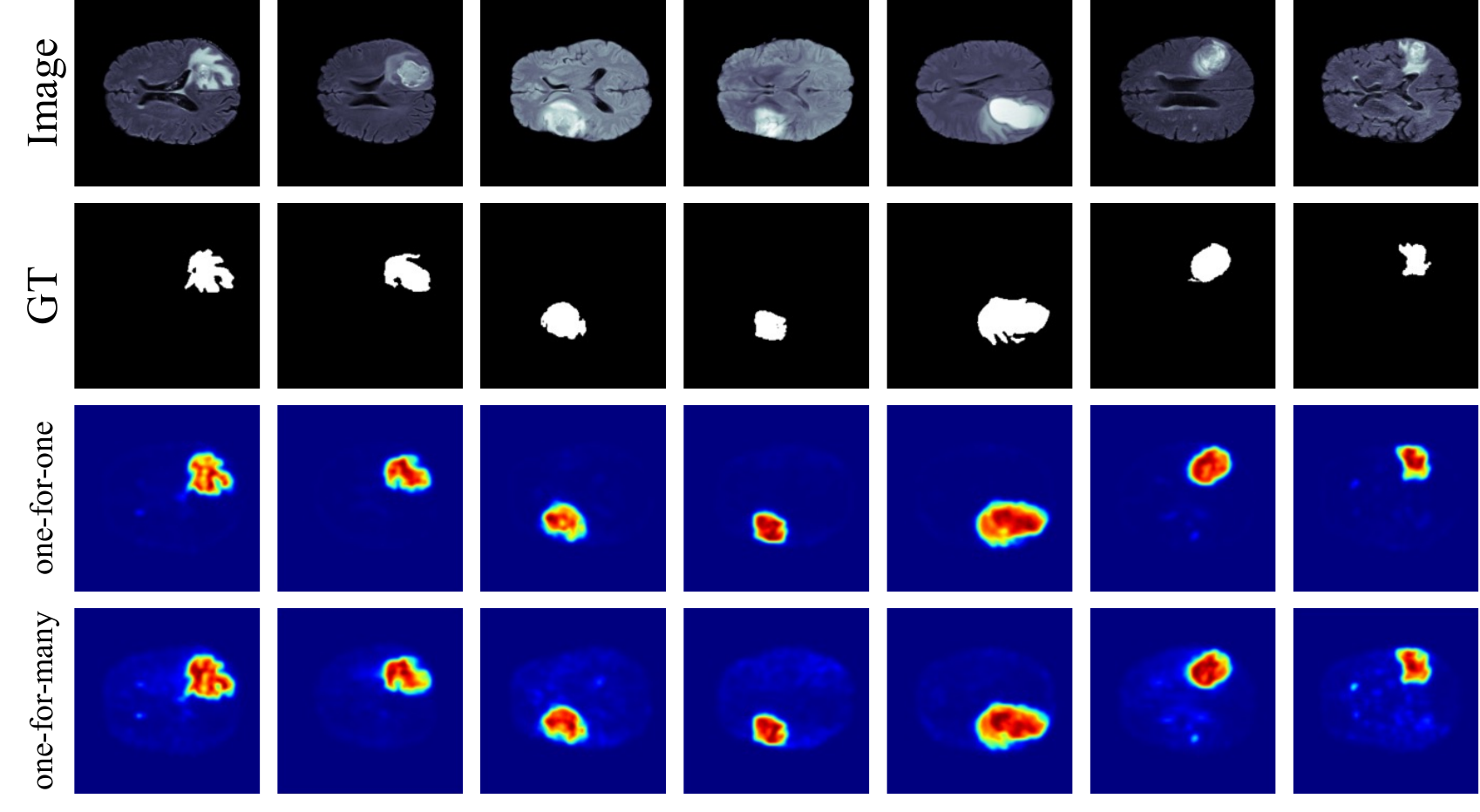}
    \caption{Qualitative anomaly localization on \textbf{BrainMRI} under the
    \textbf{2-shot} setting. Rows: input, GT, one-for-one, one-for-many.}
    \label{fig_supp:brain_2}
\end{figure}
\begin{figure}
    \centering
    \includegraphics[width=1.0\linewidth]{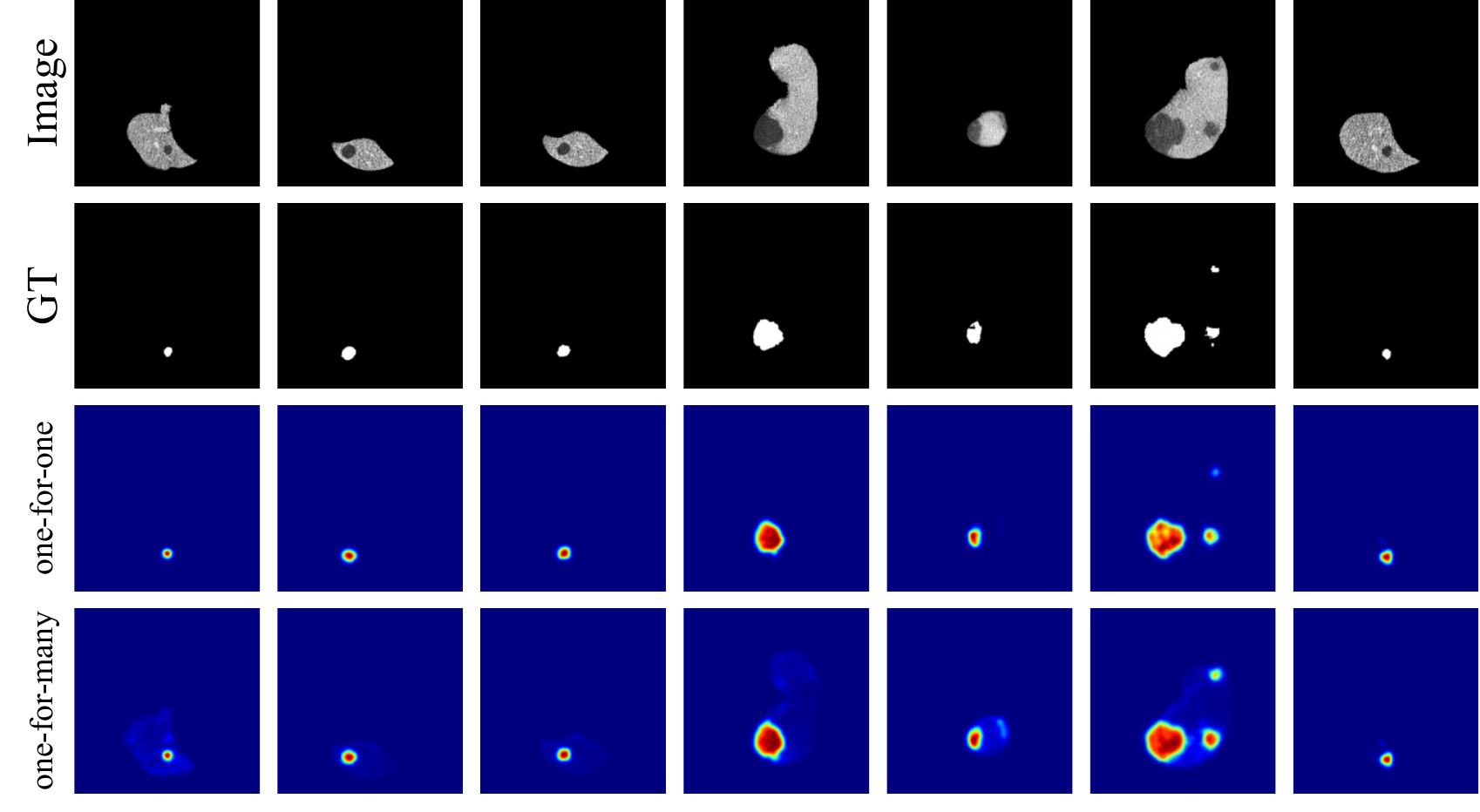}
    \caption{Qualitative anomaly localization on \textbf{Liver} under the
    \textbf{2-shot} setting. Rows: input, GT, one-for-one, one-for-many.}
    \label{fig_supp:liver_2}
\end{figure}
\begin{figure}
    \centering
    \includegraphics[width=1.0\linewidth]{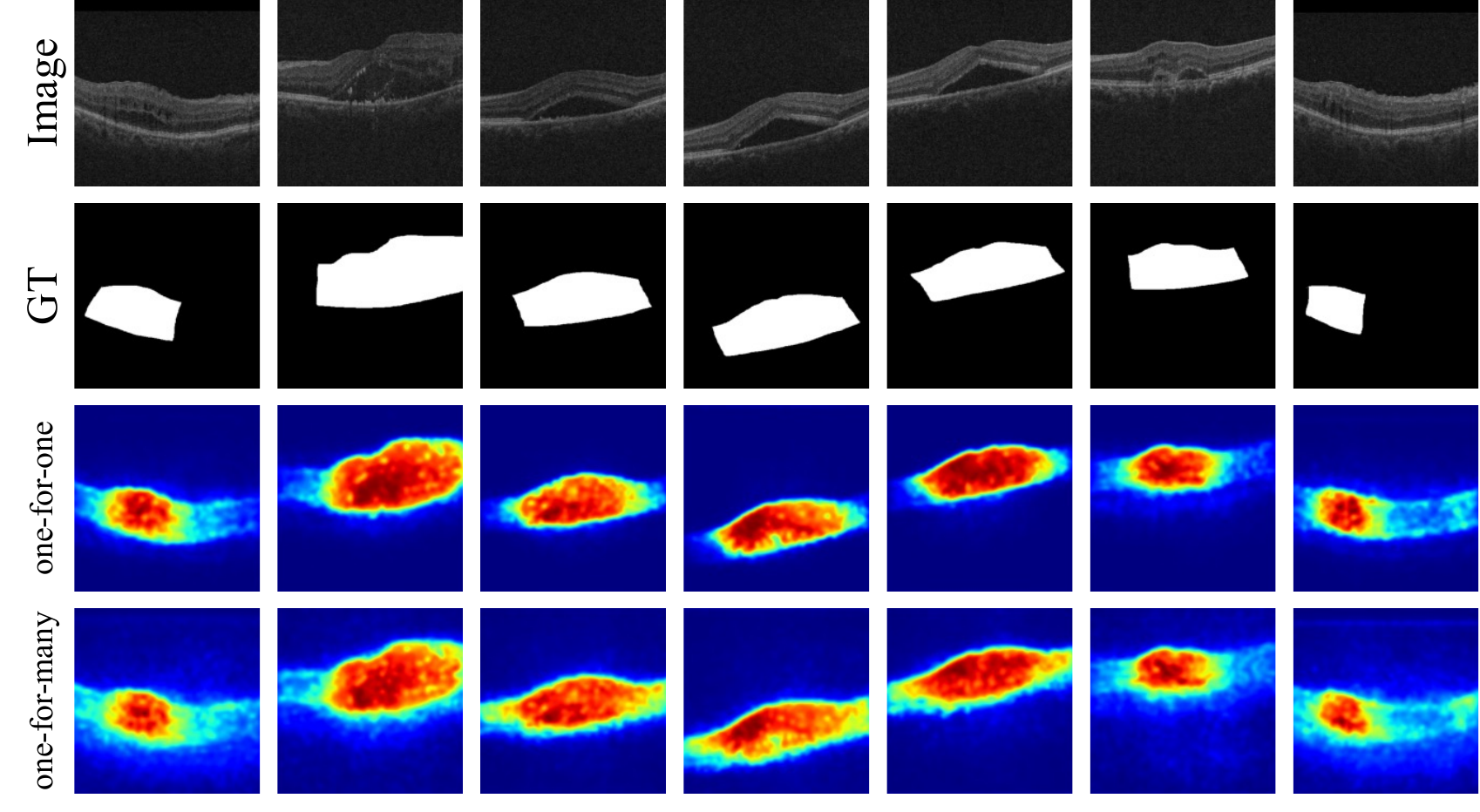}
    \caption{Qualitative anomaly localization on \textbf{RESC} under the
    \textbf{2-shot} setting. Rows: input, GT, one-for-one, one-for-many.}
    \label{fig_supp:resc_2}
\end{figure}

\begin{figure}
    \centering
    \includegraphics[width=1.0\linewidth]{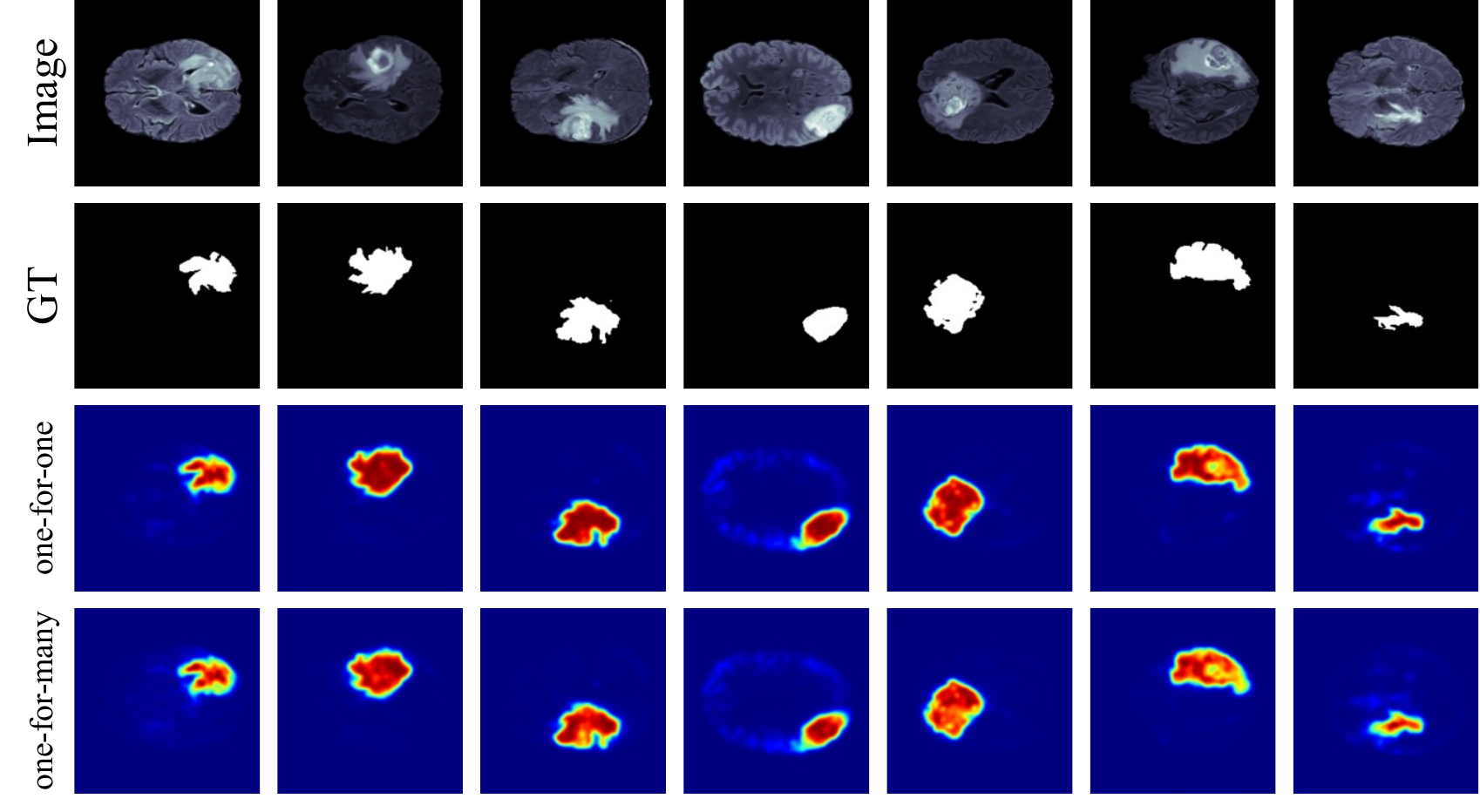}
    \caption{Qualitative anomaly localization on \textbf{BrainMRI} under the
    \textbf{4-shot} setting. Rows: input, GT, one-for-one, one-for-many.}
    \label{fig_supp:brain_4}
\end{figure}
\begin{figure}
    \centering
    \includegraphics[width=1.0\linewidth]{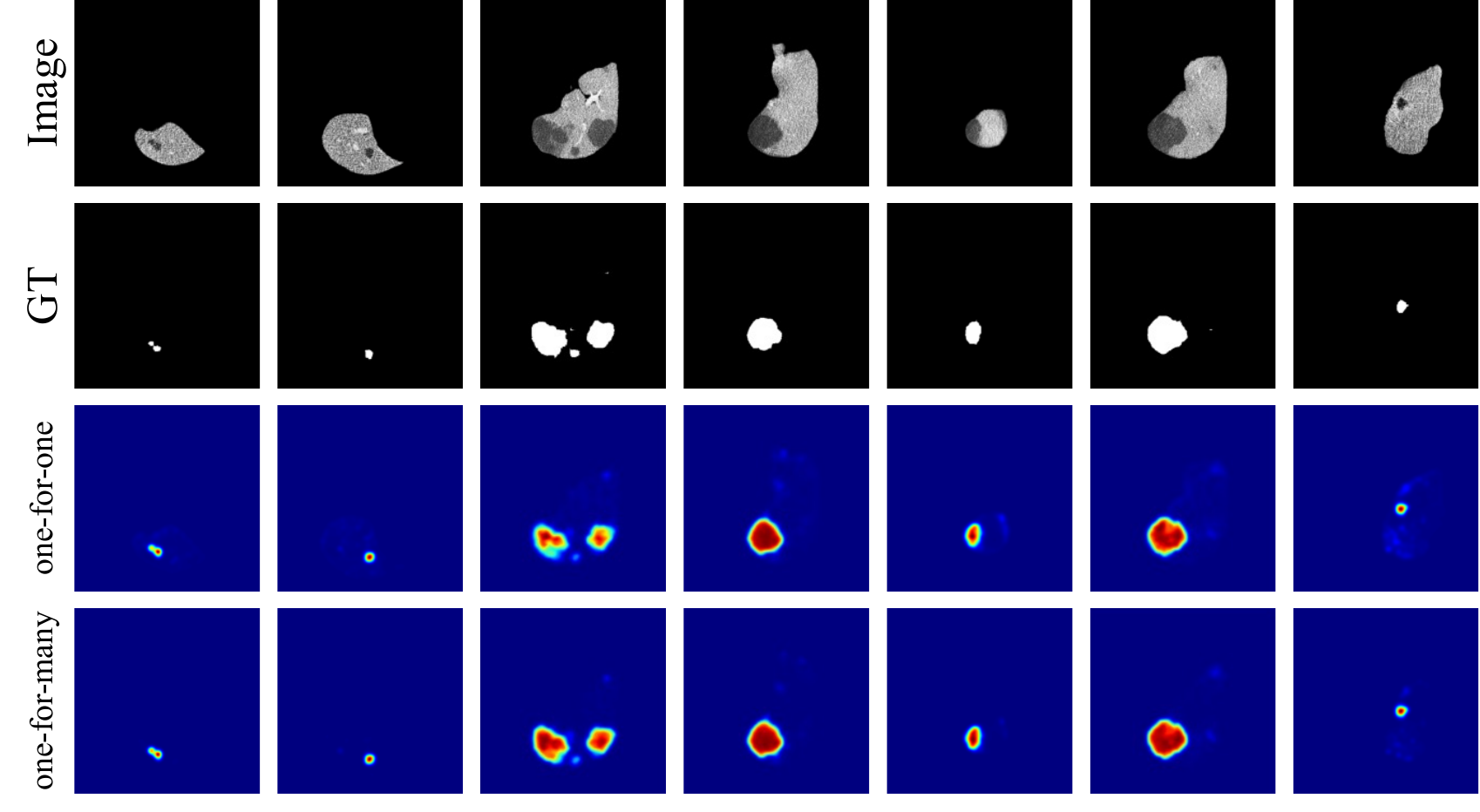}
    \caption{Qualitative anomaly localization on \textbf{Liver} under the
    \textbf{4-shot} setting. Rows: input, GT, one-for-one, one-for-many.}
    \label{fig_supp:liver_4}
\end{figure}
\begin{figure}
    \centering
    \includegraphics[width=1.0\linewidth]{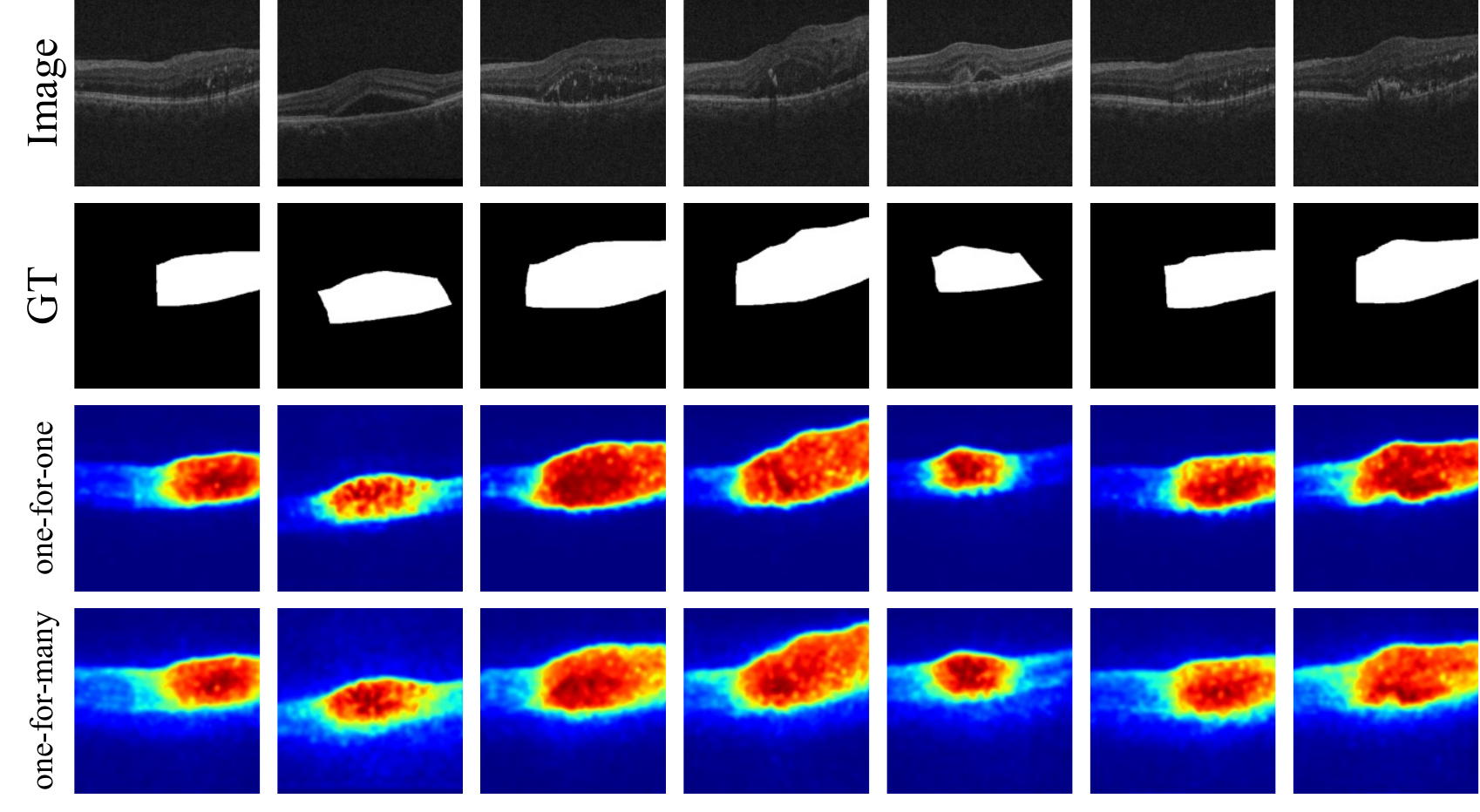}
    \caption{Qualitative anomaly localization on \textbf{RESC} under the
    \textbf{4-shot} setting. Rows: input, GT, one-for-one, one-for-many.}
    \label{fig_supp:resc_4}
\end{figure}

\end{document}